\documentclass[10pt,journal,compsoc]{IEEEtran}

\ifCLASSOPTIONcompsoc
  \usepackage[nocompress]{cite}
\else
  \usepackage{cite}
\fi

\usepackage{amsmath}
\usepackage{array}
\usepackage{svg}
\usepackage{subfig}
\usepackage{algorithm}
\usepackage{algorithmic}
\usepackage{multirow}
\usepackage{hyperref}

\usepackage{graphicx}
\usepackage{amsfonts}
\usepackage{amsmath}
\usepackage{balance}

\begin{document}

% \title{Overcoming Forgetting in Federated Learning with Selective Self Distillation}
\title{Learning Critically: Selective Self Distillation in Federated Learning on Non-IID Data}

\author{Yuting~He,
        Yiqiang~Chen,
        XiaoDong~Yang,
        Hanchao~Yu,
        Yi-Hua~Huang,
        and~Yang~Gu
\IEEEcompsocitemizethanks{\IEEEcompsocthanksitem Y.He and H.Huang are with the Beijing Key Laboratory of Mobile Computing and Pervasive Device, Institute of Computing Technology, Chinese Academy of Sciences and the University of Chinese Academy of Sciences, Beijing, China, 100190. E-mail: \{heyuting20s, huangyihua20s\}@ict.ac.cn
\IEEEcompsocthanksitem Y.Chen, X.Yang and Y.Gu are with the Beijing Key Laboratory of Mobile Computing and Pervasive Device, Institute of Computing Technology, Chinese Academy of Sciences, Beijing, China, 100190. X.Chen is also with the Shandong Academy of Intelligent Computing Technology, Jinan, China, 250101. E-mail:\{yqchen,yangxiaodong,guyang\}@ict.ac.cn
\IEEEcompsocthanksitem H.Yu is with the of Frontier Sciences and Education, Chinese Academy of Sciences, Beijing, China, 100864. E-mail:yuhanchao@ict.ac.cn\\}
\thanks{Manuscript received April 19, 2005; revised August 26, 2015.}}

\markboth{Journal of \LaTeX\ Class Files,~Vol.~14, No.~8, August~2015}%
{Shell \MakeLowercase{\textit{et al.}}: Bare Demo of IEEEtran.cls for Computer Society Journals}

\IEEEtitleabstractindextext{
\begin{abstract}
Federated learning (FL) enables multiple clients to collaboratively train a global model while keeping local data decentralized. Data heterogeneity (non-IID) across clients has imposed significant challenges to FL, which makes local models re-optimize towards their own local optima and forget the global knowledge, resulting in performance degradation and convergence slowdown. Many existing works have attempted to address the non-IID issue by adding an extra global-model-based regularizing item to the local training but without an adaption scheme, which is not efficient enough to achieve high performance with deep learning models. In this paper, we propose a Selective Self-Distillation method for Federated learning (FedSSD), which imposes adaptive constraints on the local updates by self-distilling the global model’s knowledge and selectively weighting it by evaluating the credibility at both the class and sample level. The convergence guarantee of FedSSD is theoretically analyzed and extensive experiments are conducted on three public benchmark datasets, which demonstrates that FedSSD achieves better generalization and robustness in fewer communication rounds, compared with other state-of-the-art FL methods.

\end{abstract}

}
\maketitle

\IEEEdisplaynontitleabstractindextext

\IEEEpeerreviewmaketitle

\begin{figure}[!t]
\centering
\includegraphics[width=0.9\linewidth]{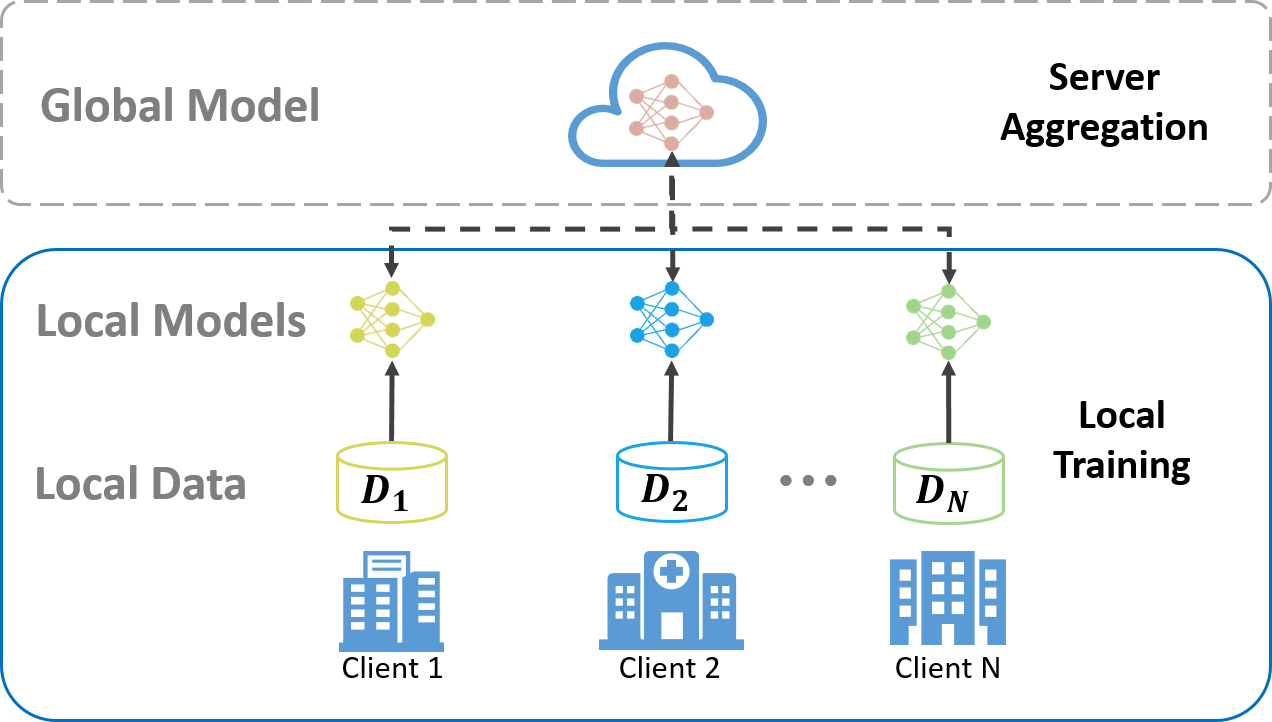}
\caption{The framework of FedAvg.}
\label{fedavg_framework}
\end{figure}

\IEEEraisesectionheading{\section{Introduction}\label{sec:introduction}}
% FL introduction
\IEEEPARstart{N}{owadays}, privacy protection has attracted increasing attention in modern society with the introduction of regulations such as the General Data Protection Regulation (GDPR) \cite{gdpr}.
The data collected by different devices or organizations cannot be gathered in a centralized server due to privacy concerns and unreliable network transmission, forming these distributed data consisting of multiple “data silos”.
Federated learning (FL) has been proposed to cope with the “data silos” dilemma, which enables clients to collaboratively train a generalized and robust model while keeping their local data decentralized. 
Most existing FL algorithms follow the procedure of FedAvg \cite{fedavg}, as shown in Fig \ref{fedavg_framework}.
In each communication round, the server sends the global model to the clients which participate in the federation.
Then, these clients locally update the models using their private data and send the optimized models back to the server. 
Finally, the server aggregates the local models to update the global model.
The above procedure is repeated until the global model converges.

% non-IID problem
One of the key challenges that distinguish Federated Learning from traditional distributed learning is the heterogeneity of data distribution across the clients, also known as non-independent and identically distributed (non-IID). The heterogeneous data not only makes the theoretical analysis difficult \cite{theory_analyze, convergence}, but also leads to performance degradation and convergence slowdown \cite{weight-divergence, li2020federated}.
Specifically, when each client trains a local model on biased data, it will be re-optimized towards the local optima and deviate from the global objective. 
This causes a drift in the client local updates, which is called "Client-drift" \cite{scaffold}. 
Moreover, the server averages the divergent local models, causing the updated global model to deviate from the global objective's optimum.

% Literature and their Limitations
To address such non-IID issue in FL, a wealth of literature has been proposed. FedProx \cite{fedprox} directly limited the local updates by adding a L2 regularization term to the local objective;
MOON \cite{moon} proposed a model-contrastive loss to control the local updates, which utilized the similarity between the global model representation and the representations of local models;
Making an analogy with Continual Learning (CL), FedCurv \cite{fedcurv} added a penalty term to the local objective to prevent the important parameters of the global model from changing too much.
However, these works regularize the local updates without adaptive adjustments to the performance of the global model, which are not efficient enough to achieve good performance for deep learning models and on more heterogeneous data.

% our main challenge
Based on the empirical observations that local models tend to overfit local data and hence forget the global knowledge. An intuitive strategy is to utilize the prediction of the global model to regularize local models, making local models preserve the knowledge that the local distributions cannot represent. 
However, unlike the teacher in traditional knowledge distillation, which is a high-capacity model capable of achieving high accuracy, the teacher (global model) in FL cannot extract feature representation well in each round and for each class, especially in the early communication rounds and for some hard classes. 
The credibility of the distilled knowledge varies with samples, classes and communications rounds.

% our method
In this paper, we propose Selective Self Distillation in FL (FedSSD), which alleviates knowledge forgetting issue by selectively self-distilling the knowledge of the global model into the local models.
We regard local models as students to critically learn the feature representations from the global model (teacher), aiming at preserving local models' global knowledge while avoiding the misleading of the global model.
More specifically, we measure the credibility of the distilled features from the global model in two levels: sample level and class level.
The class level credibility is estimated by the global model's performance on the auxiliary dataset in the server and then sent to each selected client as a credibility matrix. The sample level credibility is estimated by the global model's probability for the true class on the local samples.
During the local training, the local model selectively distills the global feature representation on local data in consideration of the two levels.
Our extensive experiments demonstrate that FedSSD can accelerate the FL training process and improve the generalization ability by alleviating catastrophic forgetting, especially on extremely heterogeneous data and for deep learning models.

% contribution
Our main contributions are summarized as follows:
\begin{itemize}

    \item We study local models' catastrophic forgetting of the global knowledge in federated learning, which causes performance degradation and convergence slowdown. We suggest that knowledge of the global model is of different credibility between samples and classes and a naive distillation strategy is not appropriate to the FL scenario.
    \item We propose a novel FL algorithm, FedSSD, to effectively mitigate the local update drift and catastrophic forgetting by selectively self-distilling the knowledge of the global model into local models.  
    \item We analyze the advantages of FedSSD by performing experiments and related visualizations that show the superiority of FedSSD in terms of classification accuracy and convergence rate on several benchmark datasets. 
\end{itemize}

Note that, this article is an extension of our earlier publication \cite{fedcad}. Our changes include method improvement and experiment enrichment. First, this article proposes a new technique for non-IID issue in FL. Our previous work ~\cite{fedcad} adaptively controls the degree of self-distillation according to the category of training samples. In addition to ~\cite{fedcad}, we analyze several aspects that need to be considered in the weight control of self-distillation in Sec.~\ref{subsec:fedssd}. Different from ~\cite{fedcad}, FedSSD applies the class-wise distillation weighting on the logits-channel level and also considers the distilling credibility on the sample level. With the new technique, we achieve a performance higher than our original submission. Second, we conduct more experiments and related analysis in Sec.~\ref{sec:experiments}. We provide additional comparison results on the cases of changing client sample ratios, changing number of local epochs and different datasets. Ablation study explores the key properties of FedSSD.

The rest of this paper is organized as follows: 
First, we review the related work in Section 2. 
Then, we introduce the motivation for our work in Section 3. 
In Section 4, we provide the detailed algorithm of our method FedSSD and mathematically analyze the convergence. 
In Section 5, extensive experimental results are reported and analyzed, before concluding in Section 6. 

%%%%%%%%%%%%%%%%%%%%%%%%%%%%%%%%%%%%%%%%%%%%%%%%%%%%%%%%%%%%%%%%%%%%%%%%%%%%%%
\section{Related Work}
\subsection{Federated Learning}
Federated learning (FL) is first proposed by \cite{fedavg} to address privacy concerns as a distributed machine learning paradigm. A key challenge in federated learning is that the data are usually non-identically distributed (non-IID) across different clients in the real world. Recently, considerable studies have been proposed to address the non-IID problem, including improvements on the local training phase and server aggregation phase. Our work focus on the first one. 

In the literature on improving the local training phase to tackle client drift, regularization terms are commonly used to impose constraints on updating the local model.
FedProx \cite{fedprox} adding a proximal term in the local objective to limit the L2 distance of the local model and current global model. Although the proximal term pulls the local model backward closer to the global model and can slightly mitigate objective inconsistency, it slows down the convergence rate \cite{fednova}.
SCAFFOLD \cite{scaffold} introduces control variates to estimate the update direction of the local model and global model. Then, the difference between these two variates is used to correct the client drift. As each client needs to send the updated model and the updated control variates, the communication burden of SCAFFOLD is doubled compared with FedAvg. 
Recent work MOON \cite{moon} empirically showed that FedProx and SCAFFOLD fail to achieve good performance on image datasets with deep neural networks. To address these issues, MOON utilizes the similarities between feature representations of the local model and global model to perform contrastive learning in the model level. 
However, we find that not all the representations learned by the global model are useful for local training. 
FedCurv \cite{fedcurv} is motivated by Continual Learning(CL), using Elastic Weight Consolidation \cite{ewc} to alleviate the forgetting issue in FL. It can prevent the important parameters of the global model from changing too much by adding a penalty term to the local objective. However, it estimates parameter importance by the diagonal of the empirical Fisher Information Matrix in the clients and sends them to the server, which brings 2.5 times communication costs compared with FedAvg.

As for the studies on improving the server aggregation phase, some works have proposed a layer-wise aggregation strategy to adapt to data heterogeneity, applying Bayesian nonparametric to match and average the parameters. For instance, instead of averaging the parameters weight-wise without considering the meaning of each parameter, PFNM \cite{PFNM} and FedMA \cite{fedma} use the Beta-Bernoulli Process for matching parameters. Specifically, FedMA is an improved version of PFNM which extends the matching strategy from fully connected layers to CNNs and LSTMs. Another line of work tries to adjust the aggregation weights, IDA \cite{inverse} calculates the inverse distance to re-weight aggregation and FedNova \cite{fednova} use normalized local model updates when averaging. These methods are orthogonal to the above methodologies which improve in the local training phase and can be coupled with each other.

Recently, personalized federated learning has attracted significant interest from researchers \cite{deng2020adaptive, chen2021fedhealth, huang2021personalized}, which tries to train personalized local models for each client. In this paper, we study conventional federated learning, with the goal of training a single generalized and robust model for all clients. 

\subsection{Knowledge Distillation}
Knowledge Distillation (KD) is proposed to transfer knowledge from a large teacher model to a small student model \cite{HintonVD15}, which is widely used for model compression \cite{MiniLM, SunCGL19} as well as to reduce the generalization errors in teacher models (i.e., self-distillation) \cite{zhang2019your, yun2020regularizing}. 
The knowledge to be distilled is not only the soften softmax probability of the teacher \cite{HintonVD15}, but also the hidden feature vector of the teacher's penultimate layer output \cite{Hints, heo2019comprehensive, kim2018paraphrasing}. Recently, some studies theoretically demonstrate the reasons for the superiority of KD, understanding how distillation is beneficial to the student network and when distillation helps \cite{tang2020understanding, Zhou2021Rethinking, DaoKSM21}.
Several studies investigated the effects of a noisy teacher. The idea of soften \cite{lukasik2021teacher} and selective \cite{wang2021selective} distillation have been proposed to avoid the harmful distilling caused by the inaccuracy of the teacher model in class level and sample level, respectively. \cite{wei2019online} and \cite{tan2019multilingual} do it in checkpoint level and model level.
Different from prior work, we consider the credibility of knowledge more comprehensively by taking all these factors into account and further going to the class level. 
\cite{kim2021comparing} theoretically and empirically investigate Mean Squared Error (MSE) loss between the student's logits and teacher's logits better than the Kullback-Leibler divergence (KL) loss between the softened softmax probability of student and teacher. Based on these findings, we proposed weighted MSE loss between the student's logits and teacher's logits in sample level and class level.

\subsection{Knowledge Distillation in FL}
Knowledge Distillation in federated learning has recently emerged as an effective approach to tracking data heterogeneity. Numerous related works study ensemble distillation, i.e. distilling the knowledge from the ensemble of teachers (local models) to a student (global model). 
In Federated distillation (FD) \cite{jeong2018communication, seo2020}, clients share the model output parameters (logits) as opposed to the model parameters (weights or gradients) to reduce the communication costs. Then, the averaged logits are used to regularize local training. FedMD \cite{li2019fedmd} and Cronus \cite{chang2019cronus} use public dataset to get the averaged logits per sample. FedDF \cite{FedDF} use an unlabeled dataset in the server to aggregate knowledge from all received local model. Furthermore, the above methods can deal with model heterogeneity and each client can design a unique model.  

Instead of treating the ensembles of local models as teachers and transferring the knowledge into the global model, FedNTD and FEDGKD \cite{lee2021preservation, fedgkd} regard the global model as a teacher and self-distill the global model's prediction during the local training phase to preserve global knowledge. However, the above-mentioned methods do not consider the global model as a teacher in FL cannot extract feature representation well in each round and for each class. To this end, our previous method FedCAD \cite{fedcad} adaptively controls the degree of self-distillation according to the according to the category of training samples. 

%%%%%%%%%%%%%%%%%%%%%%%%%%%%%%%%%%%%%%%%%%%%%%%%%%%%%%%%%%%%%%%%%%%%%%%%%%%%%%
\section{Motivation}
Under the non-IID data scenarios, each local data distribution will not be representative of the overall global data distribution. In the local training phase, the local model steps towards the local optima and tends to forget the global knowledge, which has been verified in \cite{xu2022acceleration}. Moreover, \cite{moon} verifies that the aggregated global model extracts better feature representations than the local models. In this paper, we further verify the catastrophic forgetting issue from another perspective and observe that the global model learns different representations for different classes by an experimental study.

\subsection{Forgetting the Global Knowledge}
\begin{figure}[thp]
\centering
\subfloat[]{\includegraphics[width=0.48\linewidth]{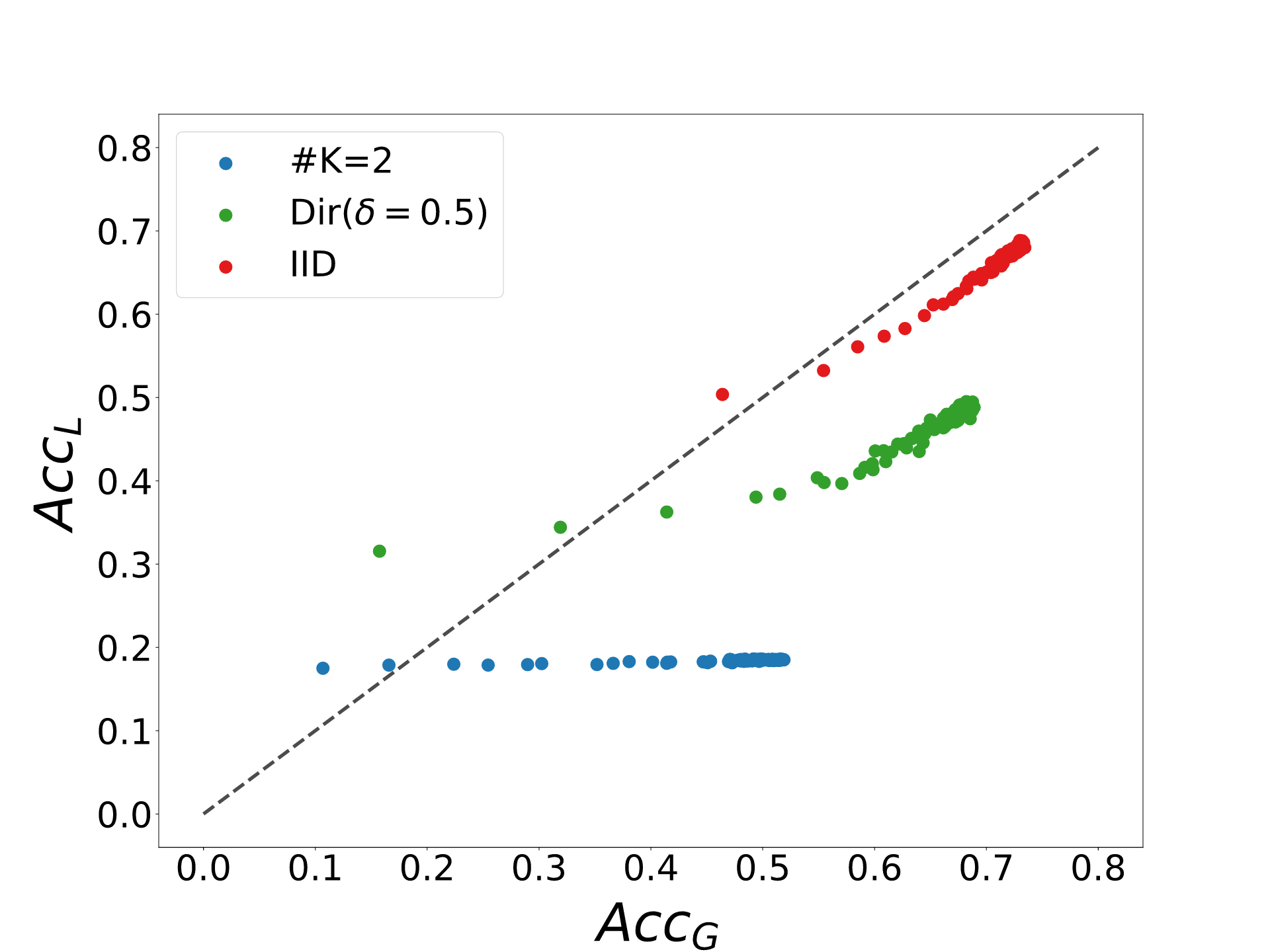}
\label{local_global_acc}}
\hfil
\subfloat[]{\includegraphics[width=0.48\linewidth]{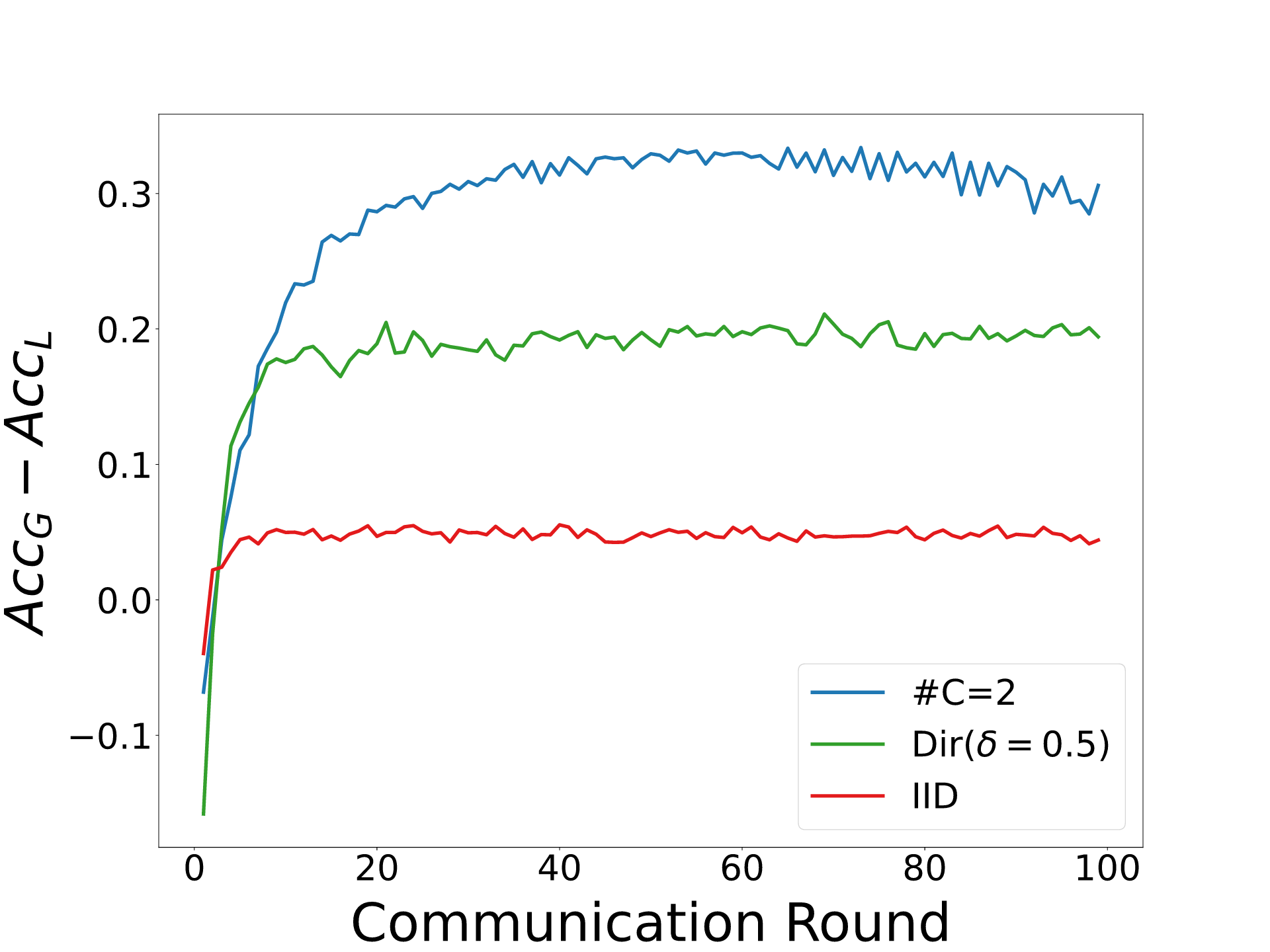}
\label{distance_acc}}
\caption{The catastrophic forgetting issue on non-IID CIFAR10. Here, $Acc_G$ and $Acc_L$ denote the global test accuracy and the average local test accuracy, respectively.}
\label{motivation_1}
\end{figure}

We compare the status of FL learning with data distributed to clients of different heterogeneity levels to illustrate the existence of catastrophic forgetting.
We choose two data partitioning strategies $\#K=k$ and $Dir(\delta=0.5)$ (see Section 5.1 for the detailed settings) on CIFAR10 to simulate the non-IID data scenario. In each round $t$, we record the performance of the global model on the uniform test dataset $Acc^{t}_G$ and the average test accuracy of the updated local models $Acc^t_L$.
As shown in Fig. \ref{local_global_acc}, we empirically observe that the values of $Acc^{t}_G$ are significantly larger than $Acc^t_L$ (a dot in the figure means one round) in the most rounds, indicating that part of the global knowledge has been forgotten during the local training phase. On the contrary, when the values of $Acc^{t}_G$ smaller than $Acc^t_L$, indicating the local models learn new knowledge from the local data while preserving the global knowledge. It is analogous to the stability-plasticity dilemma in continual learning, where the learning methods must strike a balance between retaining knowledge from previous tasks and learning new knowledge for the current task \cite{mermillod2013stability}. The forgetting issue causes the waste of learned knowledge, which reduces the learning efficiency and results in a large performance degradation. 
Furthermore, we show the difference $Acc^{t}_G - Acc^t_L$ in Fig. \ref{distance_acc}, noting that the difference increases as the heterogeneity level increases. 

\begin{figure}[!t]
\centering
\subfloat[$Dir(\delta=0.5)$]{\includegraphics[width=0.48\linewidth]{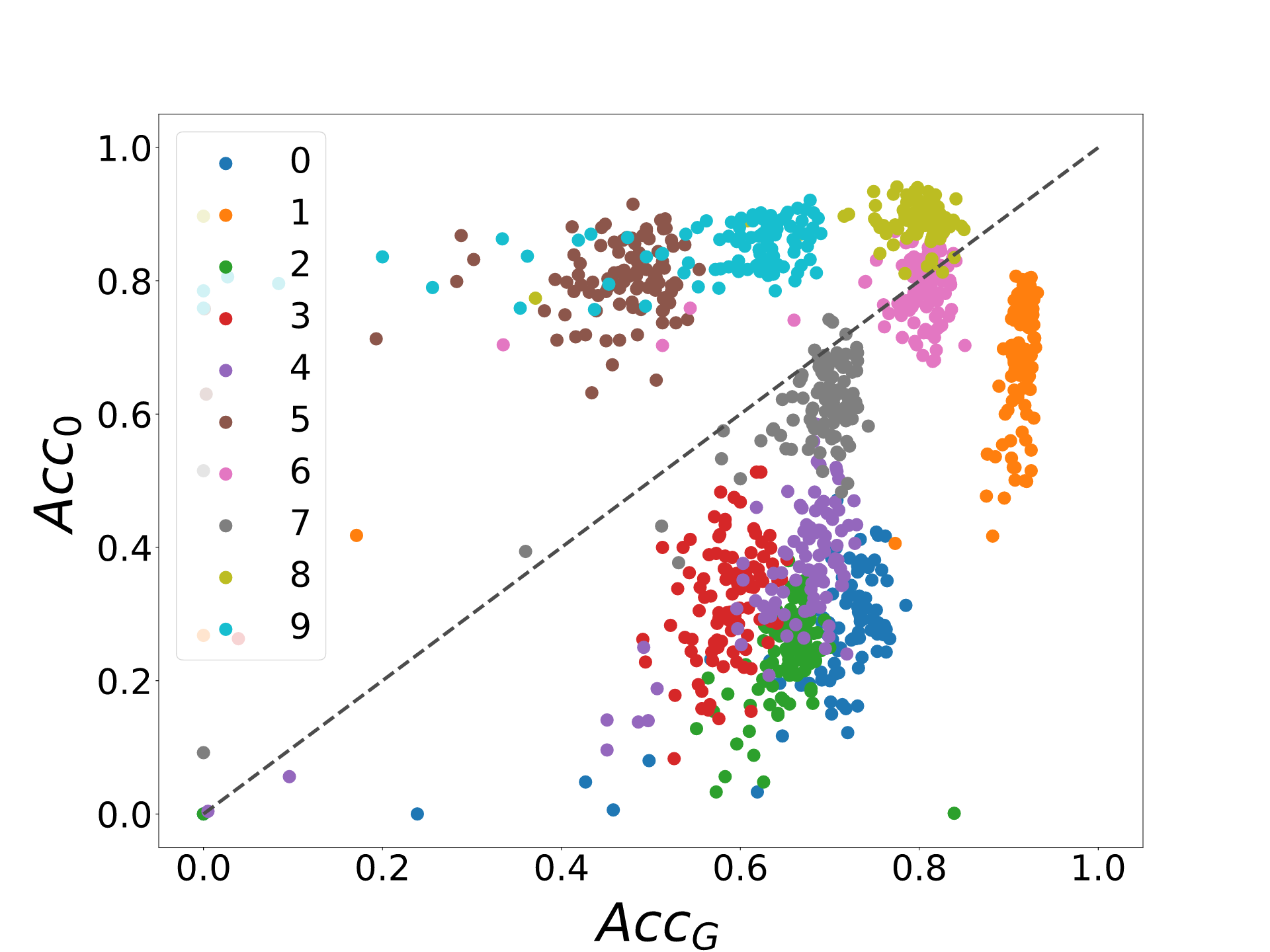}
\label{class_wise_acc_avg_dir}}
\hfil
\subfloat[IID]{\includegraphics[width=0.48\linewidth]{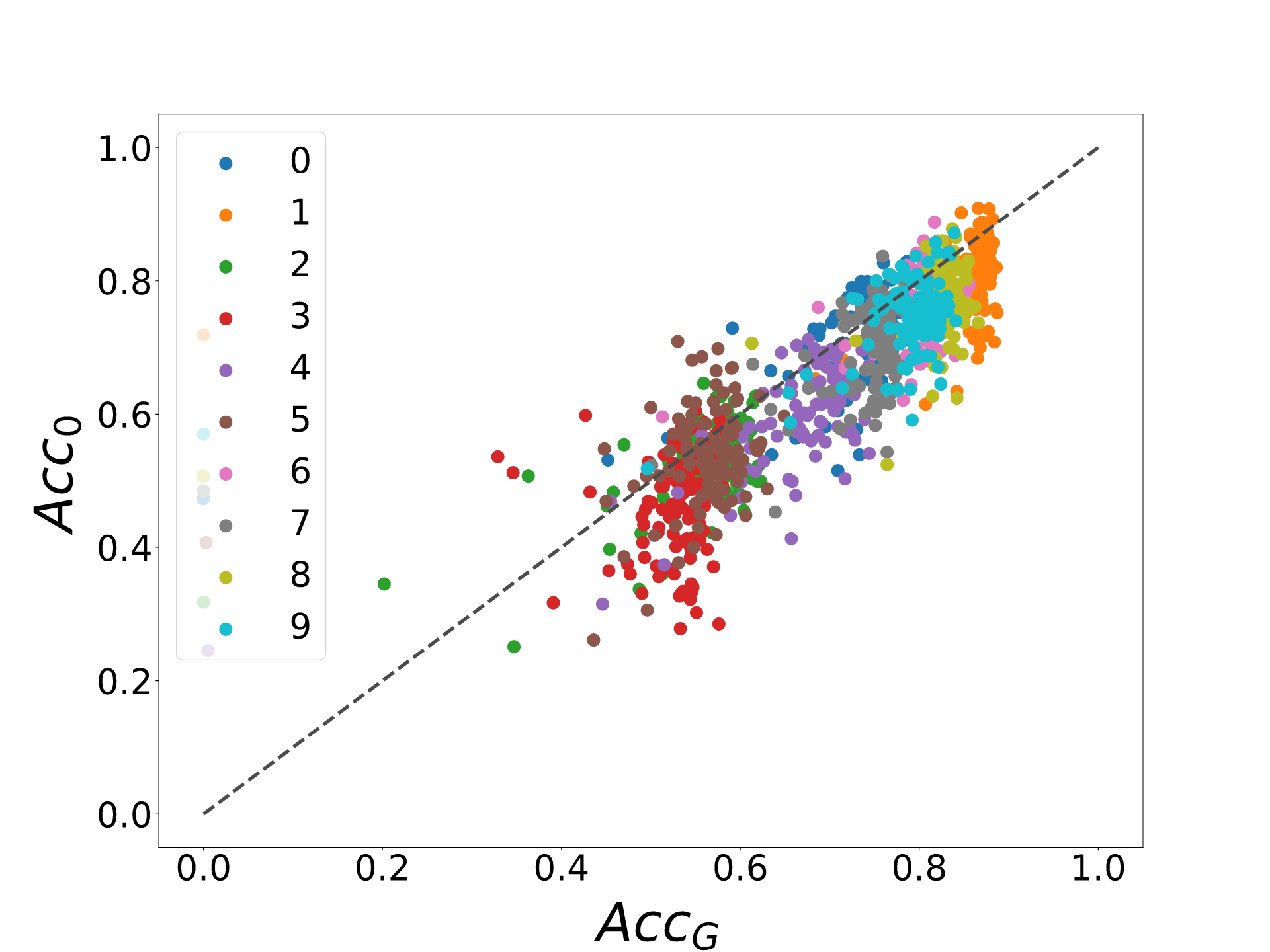}
\label{class_wise_acc_avg_dir_iid}}
\caption{The class-wise test accuracy of the global model and the local model on CIFAR10.}
\label{motivation_2}
\end{figure}

\subsection{Global Model Mis-Prediction}
Although the global model learns a better representation than the local model \cite{moon}, we conjecture that the global model perhaps cannot extract representation well at each round, on each class or sample. To verify our conjecture, we further visualize the class-wise test accuracy of the global model $Acc^{t}_G$ and the local model of "client 0" $Acc^t_0$ (whose data distribution is shown in the first column of Fig. \ref{data_distribution})  in each round. Here, we denote $Acc^t_0(k)$ as the class $k$'s accuracy of the local model $w^t_0$ in round $t$, and similarly $Acc^{t}_G(k)$ denotes the accuracy of class $k$ of the global model $w^{t}$. As shown in Fig. \ref{motivation_2}, the global model's reliability on different classes varies during the training processes. The performance of the global model in some classes is much worse than the local model, especially in the local majority class 5 and 9. 

To summarize, when the data are non-IID across clients, the local models suffer from catastrophic forgetting of the knowledge of previous training data, due to the discrepancy between local data distribution and global data distribution. Furthermore, the global model cannot perform as well as the local model on locally owned classes.

\subsection{Critically Learn from the Global Knowledge}
\begin{figure}[!t]
\centering
\subfloat[The data distribution]{\includegraphics[width=0.48\linewidth]{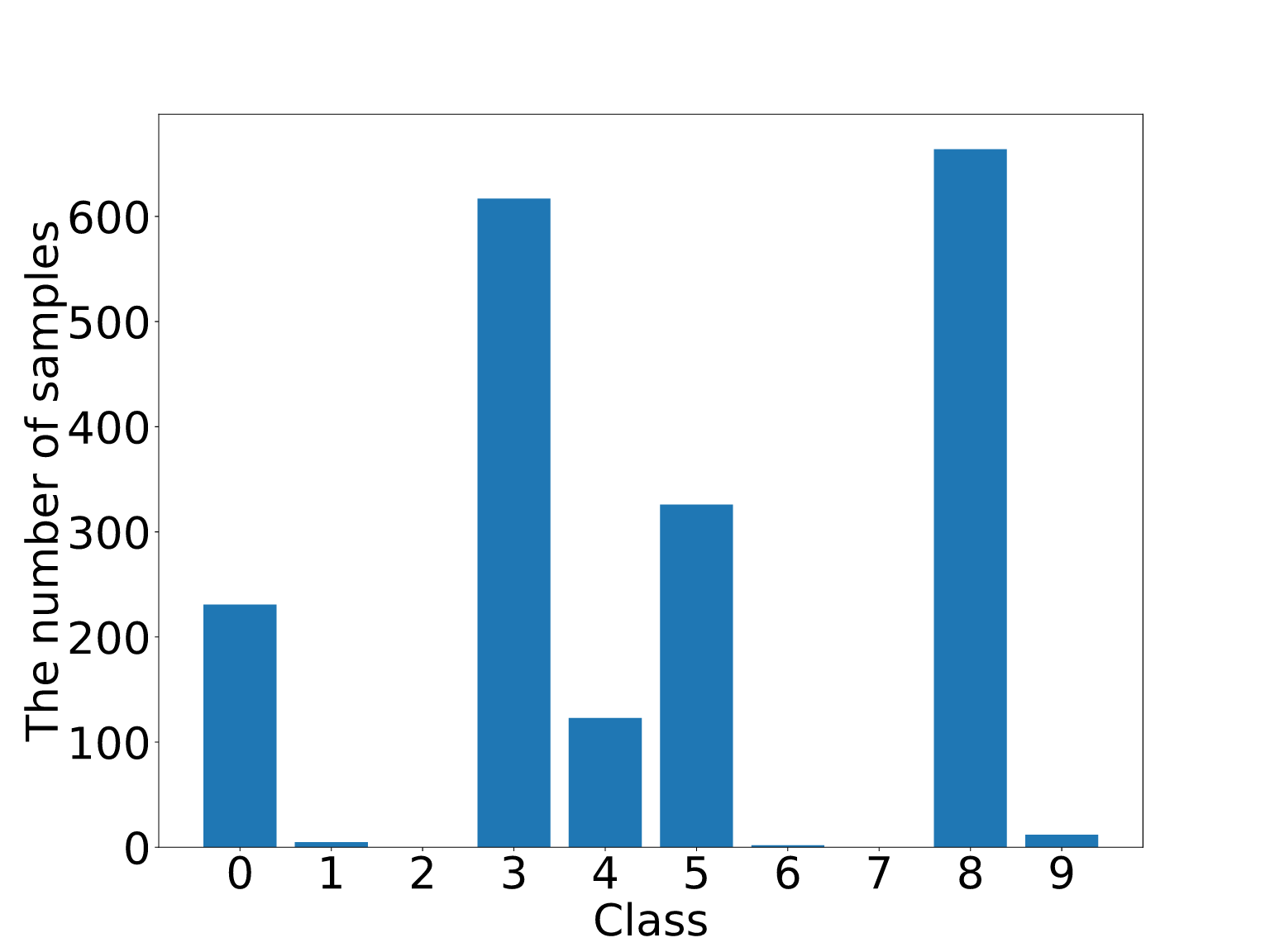}
\label{motivation_data_dir}}
\hfil
\subfloat[The global model]{\includegraphics[width=0.48\linewidth]{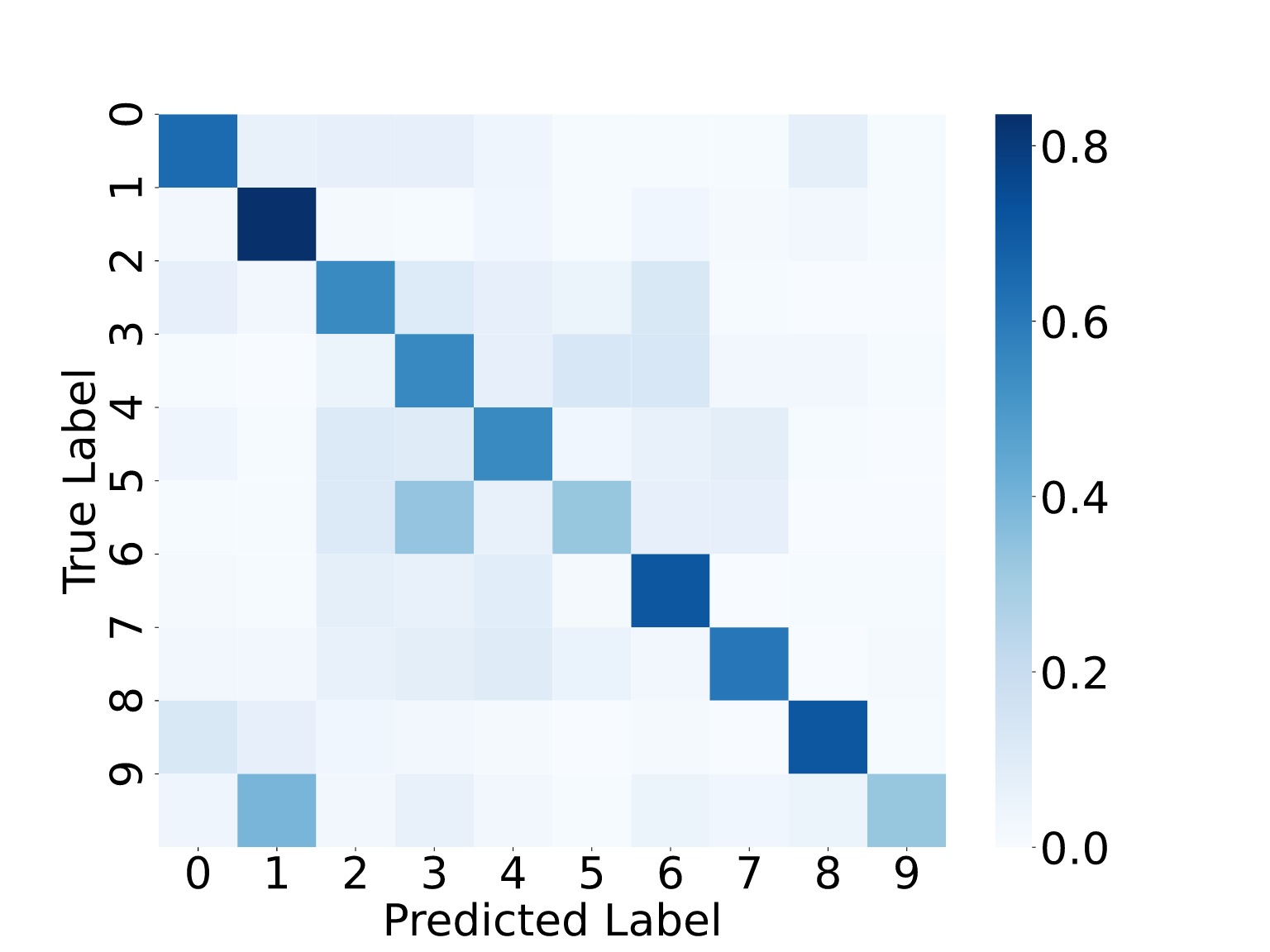}
\label{motivation_global}}
\hfil
\subfloat[FedAvg local model]{\includegraphics[width=0.48\linewidth]{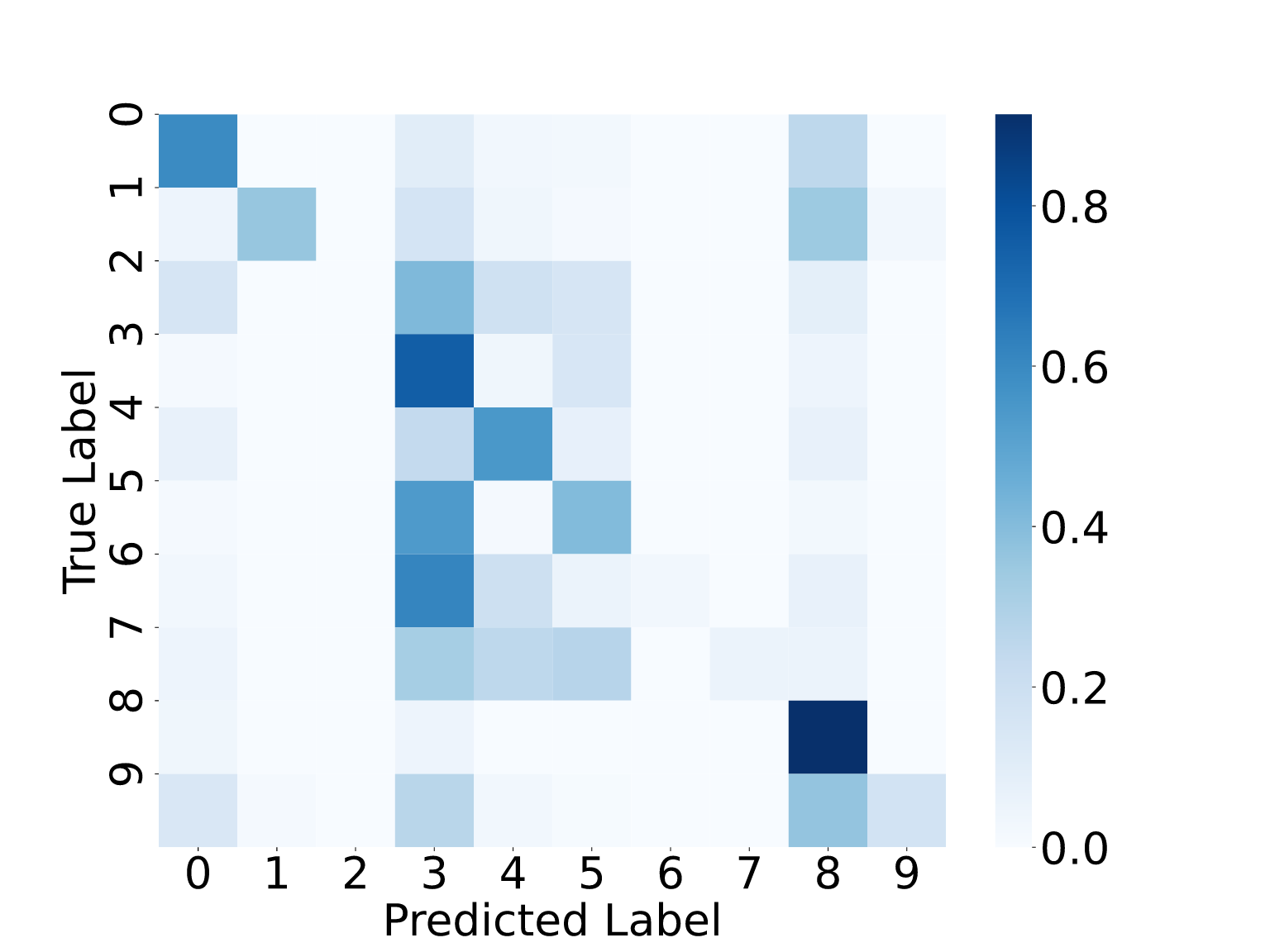}
\label{motivation_global}}
\hfil
\subfloat[FedSSD local model]{\includegraphics[width=0.48\linewidth]{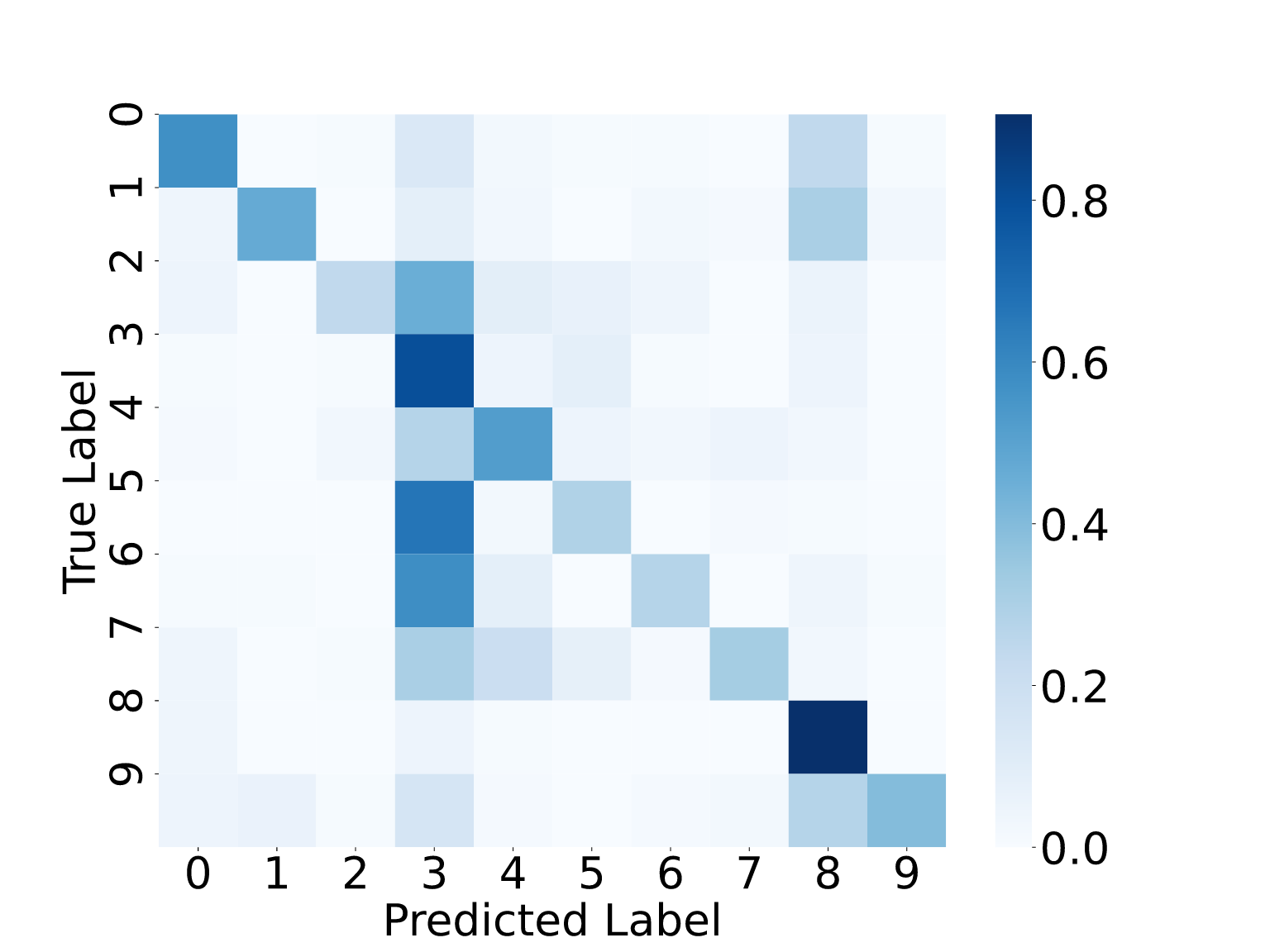}
\label{motivation_data_dir}}
\caption{Examples of confusion matrices for training on local biased data on CIFAR10.}
\label{motivation_3}
\end{figure}

Knowledge Distillation is one of the ways to keep the representations of previous data from drifting too much while learning new tasks. It is intuitive to guide the local training by the global knowledge, which is represented in the soften logits of the global model on local data. As discussed above, naively applying a constant weighted distilling loss is not a wise strategy. The reliability of the logits of the global model varies with training rounds, samples and classes. The impact of distilling should grow with the convergence of the global model. On the samples accurately predicted by the global model, the global knowledge tends to be more reliable. What's more, the global model's capability of extracting features of different classes may be different. Thus the logits of different class channels could be of different reliability. In this paper, we design a selective distilling scheme to take those factors into consideration when applying self-distilling in the local training phase. We visualize a randomly selected client's (client ID=8) confusion matrix on non-IID $Dir(\delta=0.5)$ CIFAR10 to show the superiority of FedSSD.
As shown in Fig. \ref{motivation_3}, after local training, selective distilling could not only preserves the global knowledge on the classes $\{1, 2, 6, 7\}$, but also learns local knowledge on the classes $\{0, 3, 4, 5, 8\}$. Therefore, FedSSD avoids the waste of knowledge caused by forgetting global knowledge and improves the efficiency of FL.

%%%%%%%%%%%%%%%%%%%%%%%%%%%%%%%%%%%%%%%%%%%%%%%%%%%%%%%%%%%%%%%%%%%%%%%%%%%%%%
\begin{figure*}[!t]
\centering
\includegraphics[width=0.9\linewidth]{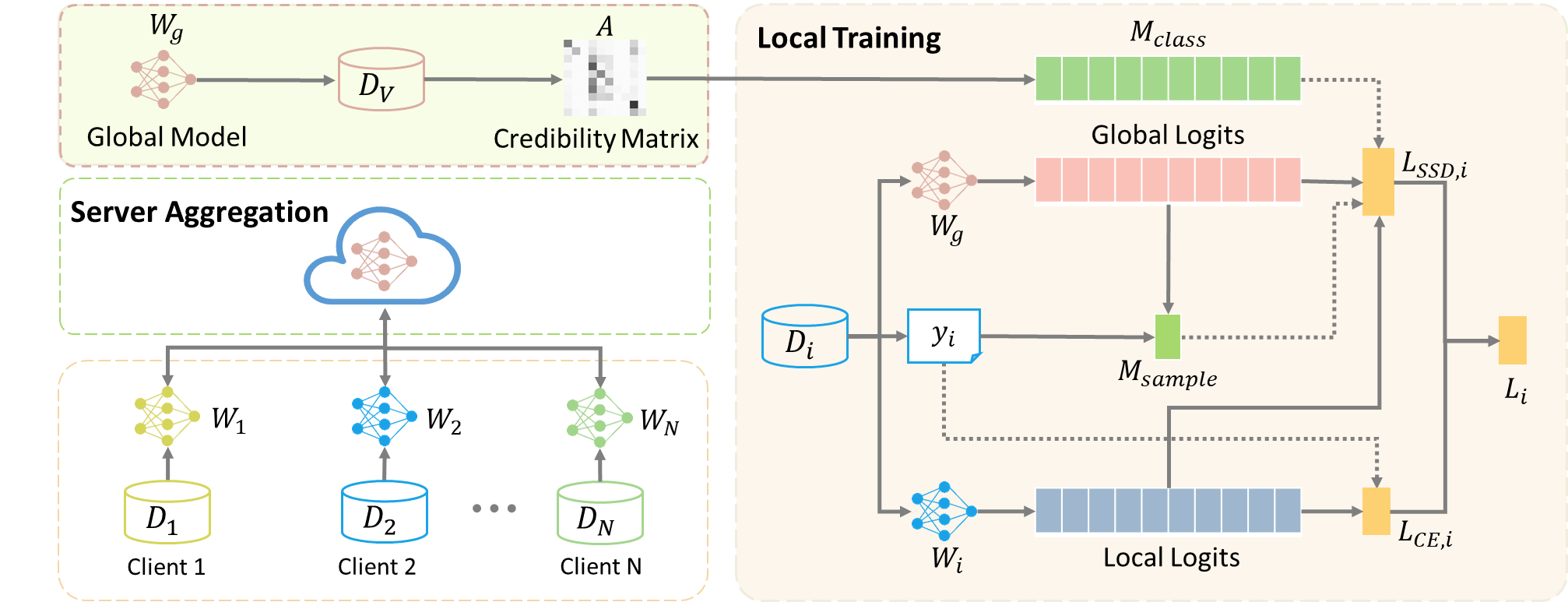}
\caption{An overview of FedSSD in the heterogeneous setting.}
\label{fedssd_framework}
\end{figure*}

\section{Methodology}
Based on the above experimental results and analysis, we propose a simple yet effective federated learning method with Selective Self Distillation (FedSSD). Since the catastrophic forgetting of the global knowledge occurs during the local update, FedSSD minimizes the discrepancy between the global model's and the local model's classifiers while learning local knowledge. 
In the following, we start with the formulation of heterogeneous federated learning and knowledge distillation. Then, we describe our critical learning strategies and the complete learning procedure. At last, we give a theoretical analysis of our method.

\subsection{Problem Formulation}
We consider there exist $N$ clients, which are all connected to a central server. Each client $i$ has a private local dataset $D_i$, with no data sharing between clients. The goal of our work is to train a generalized model $w$ to adapt to the local and global data distribution while keeping the generalization ability. More formally, federated learning can be formulated as the following optimization problem:
\begin{equation} \label{global_obj}
\min_{w} \mathcal{L}(w) = \sum^N_{i=1} q_i \mathcal{L}_i(w), \quad
q_i= |D_i| / \sum^N_{j=1} |D_j|
\end{equation}
where the global objective function $\mathcal{L}(w)$ is the weighted average of the local objectives $\mathcal{L}_i(w)$ over $N$ clients. The balancing weight $q_i$ is typically set as proportional to the sizes of the local dataset $|D_i|$. 
In general, the local objective denotes the empirical risks on possibly heterogeneous data distribution $D_i$, i.e. $\mathcal{L}_i(w):= \mathbb{E}_{(x,y) \sim D_i}[l_i(w_i;(x, y))]$, where $l_i$ measures the sample\_wise loss between the prediction of the network parameterized by $w_i$ and the ground truth label $y$ when given the input image $x$.

\subsection{Federated Selective Self distillation} \label{subsec:fedssd}

Compared to the global model, local models tend to overfit local datasets and forget global knowledge. To counteract such forgetting and raise the efficiency of federated learning, distilling knowledge from the global model to local models~\cite{fedgkd,lee2021preservation} and parameters regularization with the global model~\cite{fedprox} are proposed. However, it is not proper to distill knowledge from the global model to local models using a constant coefficient for the distilling loss term. The reliability of the predictions of the global model grows with the convergence of the global model. Distilling may mislead local models and even decrease the efficiency of federated learning at an early stage when the global model performs poorly. 

% class level:
The global model's performance in extracting features of different classes may vary due to the imbalanced data distribution of the clients. Assume that the global model is aggregated by the clients trained on skewed datasets, thus it is well-trained to extract features of the majority classes of the previously picked clients' local data and performs well on them. The output logits of the global model may be precisely estimated in channels of these classes. However, the logits in the minority classes' channels are less creditable since the aggregated local models do not learn rich features about these classes. 

% sample level:
What's more, the reliability of logits is also related to the specific training sample. In the example above, the logits on the majority class channels of the samples belonging to the majority classes are relatively more creditable on account of the accurate predictions of the global model on samples of those classes. However, it is not guaranteed that logits on these channels are still reliable for the samples of the minority classes. The global model may fail to analyze the features of the samples and predict inaccurate logits.

Traditional distillation with KL divergence~\cite{HintonVD15}, where the predictions of the global model on each class channels share the same scale of impact, is not proper for such circumstances.
The global knowledge is represented in the output logits of the global model on local data. Logits could be regarded as absolute estimations for classes. Passing through a softmax layer, logits are converted to the predicted relative probability of the samples belonging to each class.
To disentangle the predictions of the global model on different classes, we choose logits matching with L2 norm loss in distillation rather than the KL-divergence loss for aligning the probability distributions. We estimate the reliability of logits at each channel by the recall rate and the degree of inductive bias of the corresponding class on the auxiliary dataset. A logit of class channel $k$ is more reliable when the global model could fully extract the features of class $k$ and incorporate them into the probability estimation, which could be reflected in a higher recall rate of class $k$. What's more, the global model should have a less inductive bias for the channel $k$, which is reflected in a less rate of mistaking other classes for $k$. The reliability of logits on each sample is measured by the predicted probability for the true class. When the global model fails to extract and analyze features of a sample, the logits are generally imprecise and should be given less weight for distilling.

We next formulate this mathematically.
In each communication round $t$, we evaluate the credibility matrix (also called as confusion matrix) of the global model on the auxiliary dataset $D_V$ in the server and send it to the online clients in current round. We define the credibility matrix $A^t \in \mathbb{R}^{K \times K}$ of the global model as follows: 
\begin{equation}
A^t = \mathcal{P}(w^t, D_V)  \label{matrix}
\end{equation}
where $\mathcal{P}$ function denotes the performance of the global model with parameters $w^t$ on the auxiliary data $D_V$ and $K$ is the number of classes. More specifically, $A_{k_1,k_2}$ denotes the probability that the global model predicts the class $k_1$ as the class $k_2$. 

Suppose client $i \in S^t$ initializes its local model $w^{t}_i$ with the global model $w^{t}_g$ and trains on the local data $D_i$. Then, it optimizes its local objective by running Stochastic Gradient Descent (SGD) for $E$ local epochs to get the updated local model $w^t_i$. 
For every input $x$, we denote the output logits vector of the global model and the local model as $z^g=f(w^t_g, x) \in \mathbb{R}^{K}$ and $z=f(w_i^t,x) \in \mathbb{R}^{K}$, respectively. Similar to \cite{kim2021comparing}, we directly computes the mean squared error (MSE) between the global model logits and the local model logits. The selective self-distillation loss is defined as follows:

\begin{equation}
\mathcal{L}_{SSD,i} = \mathbb{E}(||M\odot z^g-M\odot z||^2_2) \label{loss_ssd}
\end{equation}
where $M\in\mathbb{R}^K$ is the class-wise weights vector and $\odot$ is element-wise multiplication. As discussed above, $M$ is related to (1) sample level: the prediction performance of the global model on each sample and (2) class level: the reliability of the logits on each class channel. Suppose the sample $x$'s label is $k_2$ and the $k_{1th}$ value of $M(x)$ is determined as:

\begin{equation}
    M(x)[k_1] = M_{max}\cdot \big[M_{class}[k_1]M_{sample}(x) - 0.1\big]^+
\end{equation} 
\[
    M_{class}[k_1] = A_{k_1,k_1}(1-\max\limits_{k\neq k_1}A_{k,k_1}) \nonumber
\]
\[
    M_{sample}(x) = 1-(1-p^g(x)[k_2])^{0.5} \nonumber
\] 
where $\big[*\big]^+=\frac{(*+|*|)}{2}$; $A_{k_1, k_1}$ is the recall rate on class $k_1$; $A_{k, k_1}$ is the rate of mistaking class $k$ for $k_1$ and $p^g(x)[k_2]$ is the global model's predicted probability for the true class $k_2$. $M_{max}$ decides the upper bound of the distillation impact. $M_{class}[k_1]$ measures the credibility of the global locit on channel $k_1$. Credible distillation on the channel $k_1$ requires the sufficient feature extraction and less inference bias on class $k_1$, which can be reflected in $M_{class}[k_1]$ and $(1-\max\limits_{k\neq k_1}A_{k,k_1})$ respectively. $M_{sample}(x)$ measures the credibility of sample $x$ by mapping the global probability on the ground truth label to $[0, 1]$ increasingly.

The classification loss $\mathcal{L}_{CE}$ is the softmax cross-entropy loss between the local model probability and the one-hot true labels $y$, which is computed as follows:
\begin{equation}
\mathcal{L}_{CE,i}=\frac{1}{|D_i|}\sum\limits_{x\in D_i} \sum^K_{k=1} -y_k\log[p_k(x)]  \label{loss_c}
\end{equation}

Consequently, the overall loss is a combination of the cross-entropy loss $\mathcal{L}_{CE}$ and the self-distillation loss $\mathcal{L}_{SSD}$ as follows:
\begin{equation}
\mathcal{L}_i = \mathcal{L}_{CE,i} + \mathcal{L}_{SSD,i}  \label{loss}
\end{equation}

The framework of FedSSD is shown in Fig. \ref{fedssd_framework} and the detailed algorithm is outlined in Algorithm \ref{alg:algorithm}. In the beginning, the server initializes the parameters of the global model $w^0$ randomly. Then, it runs the outer loop in Algorithm \ref{alg:algorithm} for $T$ communication rounds. In each round t, the server evaluates the global model on the auxiliary data $D_V$ to get the credibility matrix. Then, it samples $C*N$ clients to form the sampled set of clients $S_t$ for receiving updates. Next, the server sends the model parameters $w^t$ and credibility matrix $A^t$ to the clients in $S_t$. After receiving the parameters and the credibility of the global model, each client $i \in S^t$ executes the inner for loop in Algorithm \ref{alg:algorithm}. First, the client $i$ initializes its local model $w_i^t$ with the received global model parameters $w^t$. Then, it updates the local model by minimizing its local objective function on its local dataset $D_i$ and in the meanwhile minimizing the discrepancy between the global model's and student's logits. During the local training phase, the global model is frozen and local models update $E$ epochs by the SGD algorithm. Once the server receives all updates from the clients in $S^t$, it aggregates model parameters by a weighted average of the number of samples.

\begin{algorithm}[tb]
\caption{FedSSD: Selective Self-Distillation in FL}
\label{alg:algorithm}
\textbf{Input}: $N$ clients' datasets $\{D_i\}^N_{i=1}$, auxiliary dataset at the server $D_V$, total communication rounds $T$, clients sample ratio $C$, local epochs $E$, learning rate $\eta$, minibatch size $b$.\\
\textbf{Output}: The final global model $w^T$\\
\textbf{ServerExecute:}
\begin{algorithmic}[1] %[1] enables line numbers
\STATE Initialize the global model $w^0$ in the server
\FOR{$t=0,...,T-1$}
\STATE $A^t = \mathcal{P}(w^t, D_V)$  \hfill  $\triangleright$ \textbf{Eq.\eqref{matrix}}
\STATE $S_t \gets $ Randomly sample a set of $C \times N $ clients 
\FOR{$i \in S_t$ in parallel}
\STATE ${w^{t}_i} \gets $ \textbf{ClientUpdate($i, w^t, A^t$)} 
\ENDFOR
\STATE $w^{t+1} \gets \frac{1}{|D_{S_t}|} \sum_{i \in S_t} |D_i|w^{t}_i$
\ENDFOR
\STATE \textbf{return} ${w^{T}}$
\end{algorithmic}
\textbf{ClientUpdate:($i, w^t, A^t$)}
\begin{algorithmic}[1]
\STATE $w^t_i \gets w^t$
\FOR{epoch $e = 1,2,...E$}
\FOR{batch $b=\{ x, y\} \in D_i$ }
\STATE $\mathcal{L}_{SSD,i} \gets \frac{1}{|D_i|}\sum\limits||M\odot z^g-M\odot z||^2_2$ \hfill $\triangleright$  \textbf{Eq.\eqref{loss_ssd}}
\STATE $\mathcal{L}_{CE,i} \gets CrossEntropyLoss(f(x),y) $ \hfill $\triangleright$  \textbf{Eq.\eqref{loss_c}}
\STATE $\mathcal{L}_i \gets \mathcal{L}_{CE,i} + \mathcal{L}_{SSD,i}$ \hfill $\triangleright$   \textbf{Eq.\eqref{loss}}
\STATE $w^t_i \gets w^t_i - \eta\nabla \mathcal{L}(w^t_i,b)$
\ENDFOR
\ENDFOR
\STATE \textbf{return} ${w^{t}_i}$ to the server
\end{algorithmic}
\end{algorithm}

\subsection{Convergence Analysis}
We give the convergence analysis in this section, stating the bounded dissimilarity, Lipschitz smooth assumptions about the local objective function and Lipschitz continuity assumption about logits function, which are similar to existing literature \cite{fedprox, convergence, fednova, HarmoFL}. We introduce the following notations before formally stating the convergence result. $[N]$ denotes $\{1,2,...,N\}$ for any positive integer $N$. $x,f$ represent the input data and the logits function, respectively. $\mathcal{L}$ denotes the global objective which is the weighted sum of local objectives $\mathcal{L}_i$.

\noindent \textbf{Assumption 1.}
\begin{itemize}
    \item[1] (Bounded dissimilarity) For each client and any parameter $i\in[N]$, $w \in \mathbb{R}^W$, there exists a constant $B>0$ such that $\mathbb{E}_i[||\nabla \mathcal{L}_i(w)||^2]\leq||\nabla \mathcal{L}(w)||^2B^2$. If the local objective functions are identical to each other, then we have $B^2=1$.
    
    \item[2] (L-Lipschitz smooth) For each local objective function $\mathcal{L}_i$ is Lipschitz smooth, there exists a constant $L_F>0$ such that $||\nabla \mathcal{L}_i(w)-\nabla \mathcal{L}_i(w')||\leq L_F||w-w'||$. We further assume that $\nabla^2\mathcal{L}_i(w)$ is lower bounded by a constant $L_m$ multiplied with an identity matrix.
    
    \item[3] (L-Lipschitz continuity) For each local logits function $f$ is Lipschitz continuity, there exists a constant $L_f>0$ such that $||f(w,x_i)-f(w',x_i)||<\sqrt{2L_f}||w-w'||$. 
\end{itemize}

Remarking that \cite{fedprox} also made the Assumption 1.1 and 1.2. Assumption 1.3 holds when the training data $x$ and parameter $w$ are bounded. With bounded $x$ fed, each layer of the deep network is L-Lipschitz continuous, resulting in that the logits function $f(w, x)$ is also L-Lipschitz. Based on assumption 1.3, we introduce a lemma for the rationality of the nest assumption.

\noindent \textbf{Lemma 1.} Define $\tilde{\mathcal{L}}$ as the following:
\begin{flalign*}
&\mathcal{L}(w,w^t)&& \\\nonumber
&= \mathcal{L}_{CE}(w) + \frac{1}{2|D_i|}\sum\limits^{|D_i|}_{j=1}||M\odot (f(w,x_{i,j}) - f(w^t,x_{i,j}))||^2 && \\\nonumber
& \leq \mathcal{L}_{CE}(w) + \frac{1}{2|D_i|}\sum\limits^{|D_i|}_{j=1}M_{max}^2|| f(w,x_{i,j}) - f(w^t,x_{i,j})||^2 && \\\nonumber
& \leq \mathcal{L}_{CE}(w) + \frac{1}{|D_i|}\sum\limits^{|D_i|}_{j=1}M_{max}^2L_f^2||w - w^t||^2 && \\\nonumber
&= \mathcal{L}_{CE}(w) + M_{max}^2 L_f^2||w-w^t||^2 && \\\nonumber
&=: \tilde{\mathcal{L}}(w,w^t) && \\\nonumber
\end{flalign*}
where the first inequality holds because $M$ is upper bounded by $M_{max}$ and the second inequality follows from Assumption 1.3. For any $w^{t+1}_i$ satisfies $\tilde{\mathcal{L}}(w^{t+1}_i,w^t)\leq\tilde{\mathcal{L}}(w^t,w^t)$, then we get the following relationship:
\[\mathcal{L}(w_i^{t+1},w^t)\leq \tilde{\mathcal{L}}(w_i^{t+1},w^t)\leq \tilde{\mathcal{L}}(w^t,w^t) = \mathcal{L}(w^t,w^t)\]
which implies that a solution optimizing $\tilde{\mathcal{L}}$ is also a solution optimizing $\mathcal{L}$.

\noindent \textbf{Assumption 2.} At round $t$, the optimized parameter $w^{t+1}_i$ satisfies that:
\begin{equation}
    ||\nabla \mathcal{L}_{CE,i}(w^{t+1}_i) + M_{max}L_f(w^{t+1}_i-w^t)|| \leq \eta ||\nabla \mathcal{L}_i(w^t)|| \nonumber
\end{equation}
where the learning rate $\eta\in [0,1)$.

The Assumption 2 is well-posed based on Lemma 1, which implies the intersection of the solution space for $\tilde{\mathcal{L}}$ and $\mathcal{L}$ are not empty. 

\noindent \textbf{Theorem 1.} (Non-convex FedSSD convergence). Let Assumption 1 and Assumption 2 hold. Suppose in each round $t$, a set of $|S^t|=S$ clients are chosen. If $M_{max}$, $S$ and $\eta$ are chosen such that
\begin{flalign*}
1. \quad & \rho =(\frac{1}{M_{max}L_f} - \frac{\eta B}{M_{max}L_f} - \frac{B(1+\eta)\sqrt{2}}{(M_{max}L_f+L_m)\sqrt{S}} && \\\nonumber
         &- \frac{L_FB(1+\eta)}{(M_{max}L_f+L_m)M_{max}L_f} - \frac{L_F(1+\eta)^2B^2}{2(M_{max}L_f+L_m)^2} && \\\nonumber
         &- \frac{L_FB^2(1+\eta)^2}{(M_{max}L_f+L_m)^2 S}(2\sqrt{2S}+2)) > 0 && \\\nonumber
\end{flalign*}
\begin{flalign*}
2. \quad & M_{max}L_f+L_m>0 &&\nonumber
\end{flalign*}
Then we have the following expected decrease in the global objective:
$$\mathbb{E}_{S_t}[\mathcal{L}(w^{t+1})] \leq \mathcal{L}(w^t)-\rho||\nabla \mathcal{L}(w^t)||^2$$
The theorem derives from the Theorem 4. in ~\cite{fedprox}.
Thus, convergence can be guaranteed when there is a certain expected one-round decrease, which can be achieved by choosing appropriate $M_{max}$ and $\eta$.

%%%%%%%%%%%%%%%%%%%%%%%%%%%%%%%%% Data distribution %%%%%%%%%%%%%%%%%%%%%%%%%%%%%%%%%%%%%%%%%

\begin{figure*}[t]
\centering
\subfloat[CIFAR10]{\includegraphics[width=0.3\linewidth]{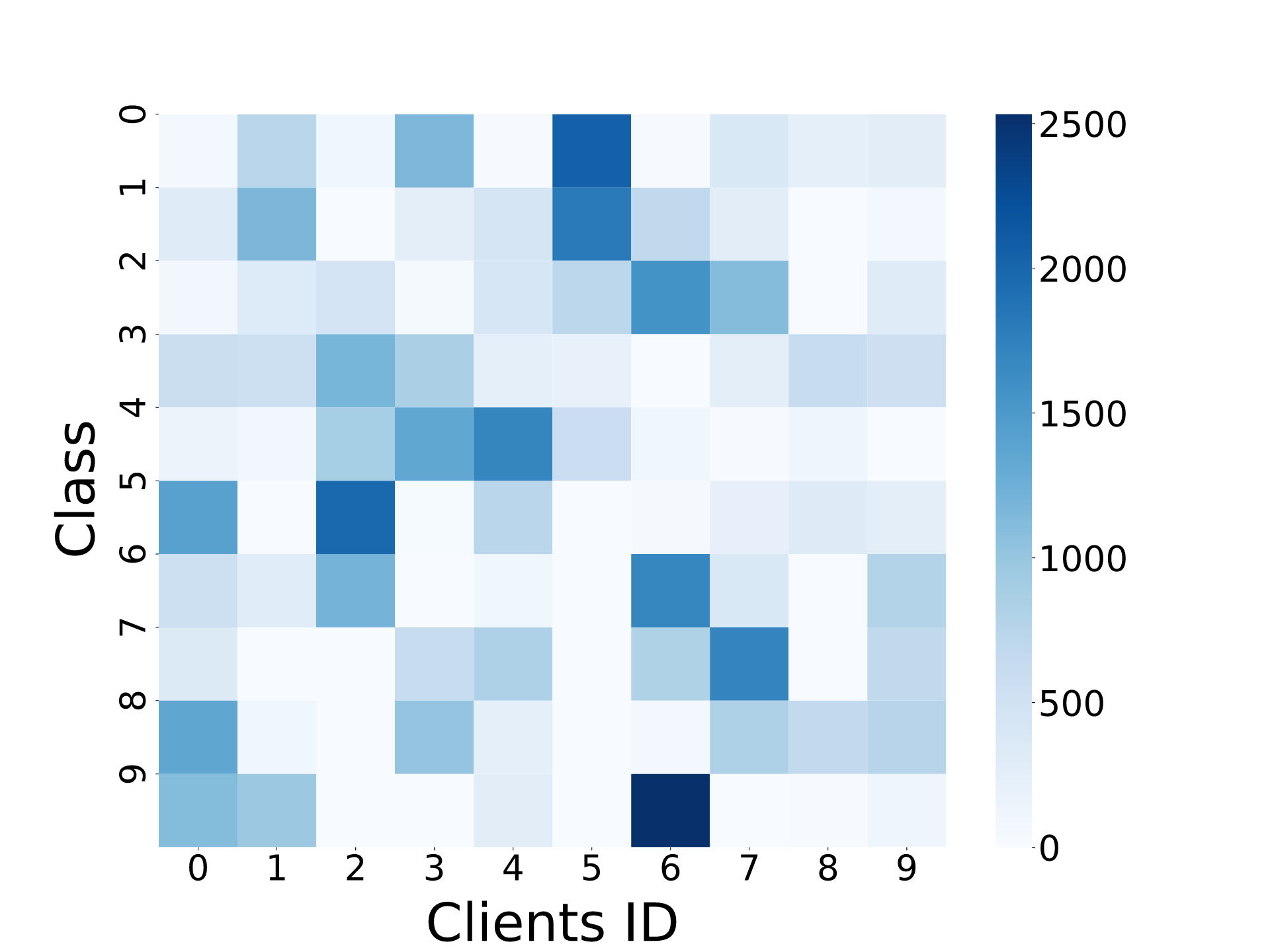}
\label{dir05}}
\hfil
\subfloat[CIFAR100]{\includegraphics[width=0.3\linewidth]{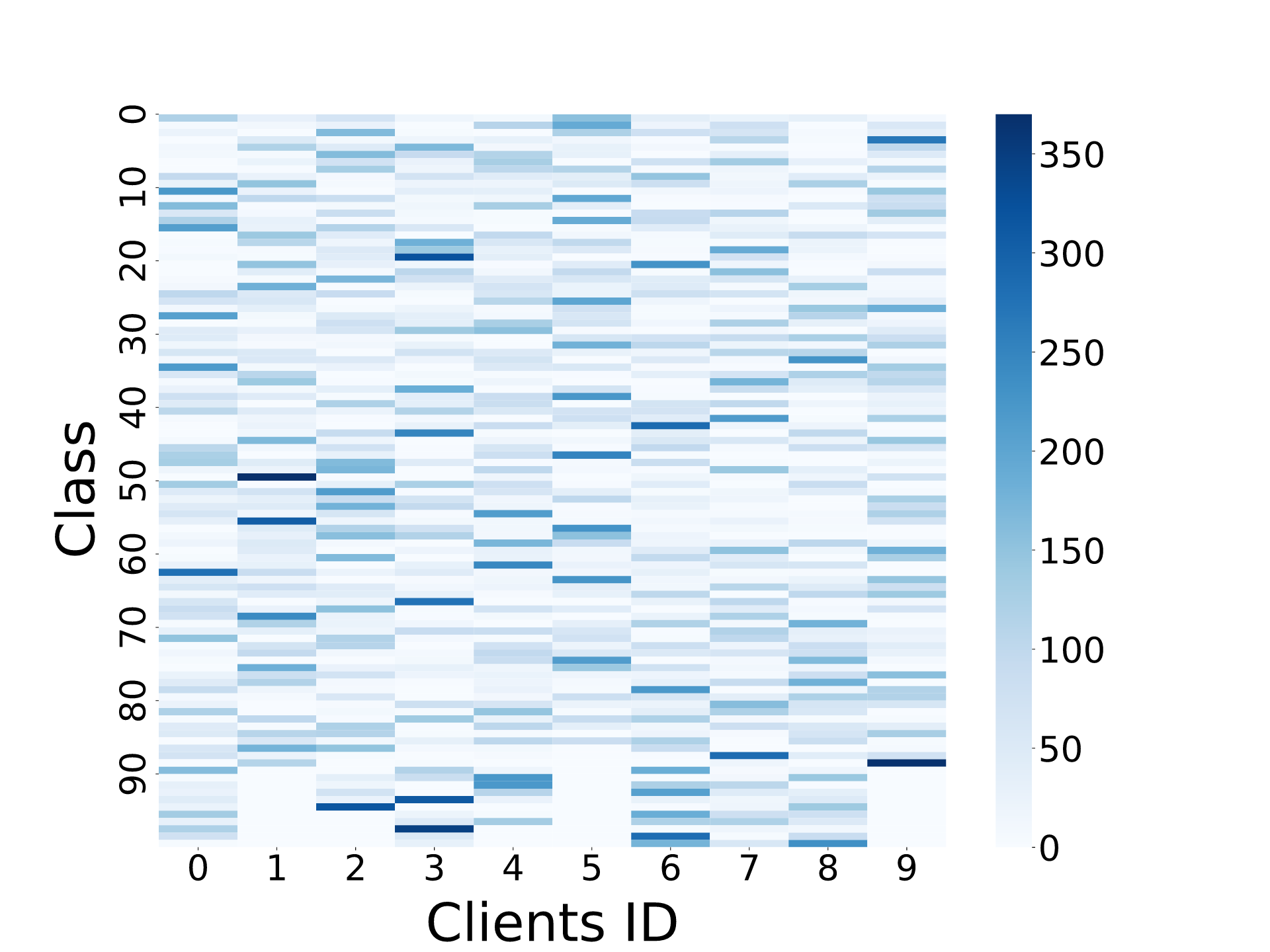}
\label{dir10}}
\hfil
\subfloat[TinyImageNet]{\includegraphics[width=0.3\linewidth]{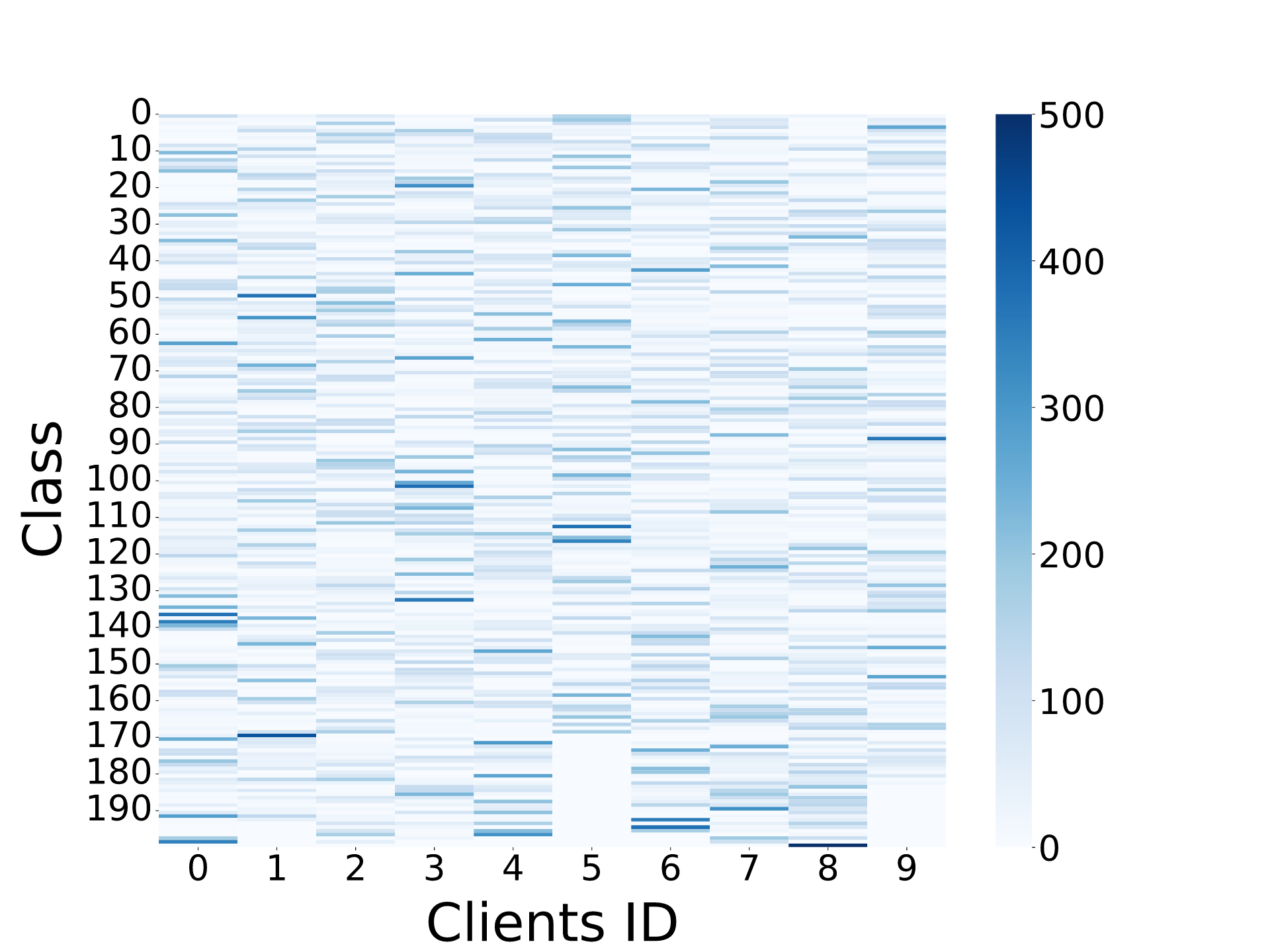}
\label{dir10}}
\caption{Visualization of statistical heterogeneity among clients, where the x-axis indicates client IDs, the y-axis indicates class labels and each rectangle indicates the number of samples held by one client
for each particular class.}
\label{data_distribution}
\end{figure*}
%%%%%%%%%%%%%%%%%%%%%%%%%%%%%%%%% Data distribution %%%%%%%%%%%%%%%%%%%%%%%%%%%%%%%%%%%%%%

%%%%%%%%%%%%%%%%%%%%%%%%%%%%%%%%%  SOTA %%%%%%%%%%%%%%%%%%%%%%%%%%%%%%%%%%%%%%%%%
\begin{figure*}[t]
	\centering
	\subfloat[CIFAR10]{
	\begin{minipage}[b]{0.3\linewidth}
	\includegraphics[width=1\linewidth]{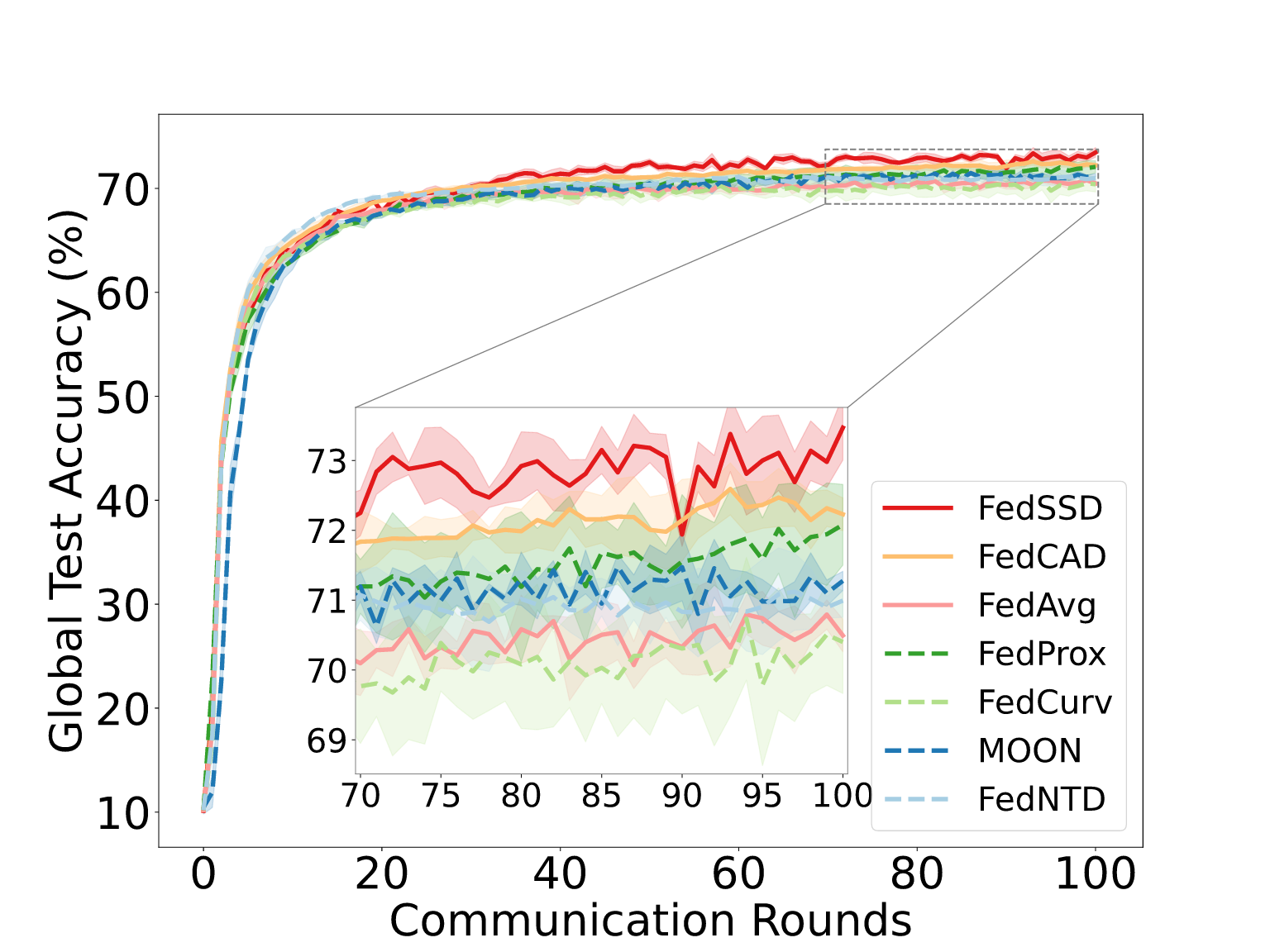}\\
		\includegraphics[width=1\linewidth]{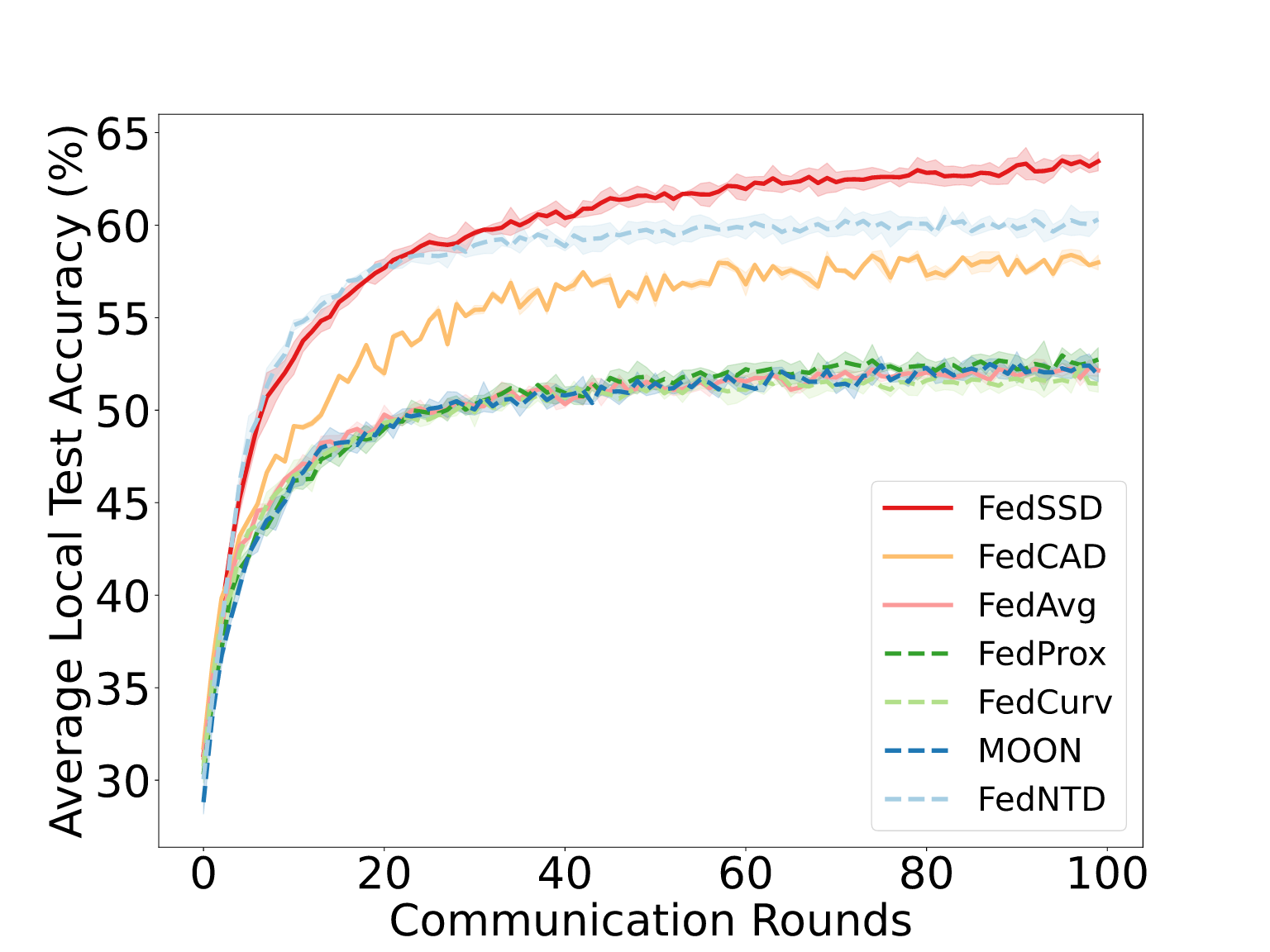}\\

	\end{minipage}
}
	\subfloat[CIFAR100]{
	\begin{minipage}[b]{0.3\linewidth}
	\includegraphics[width=1\linewidth]{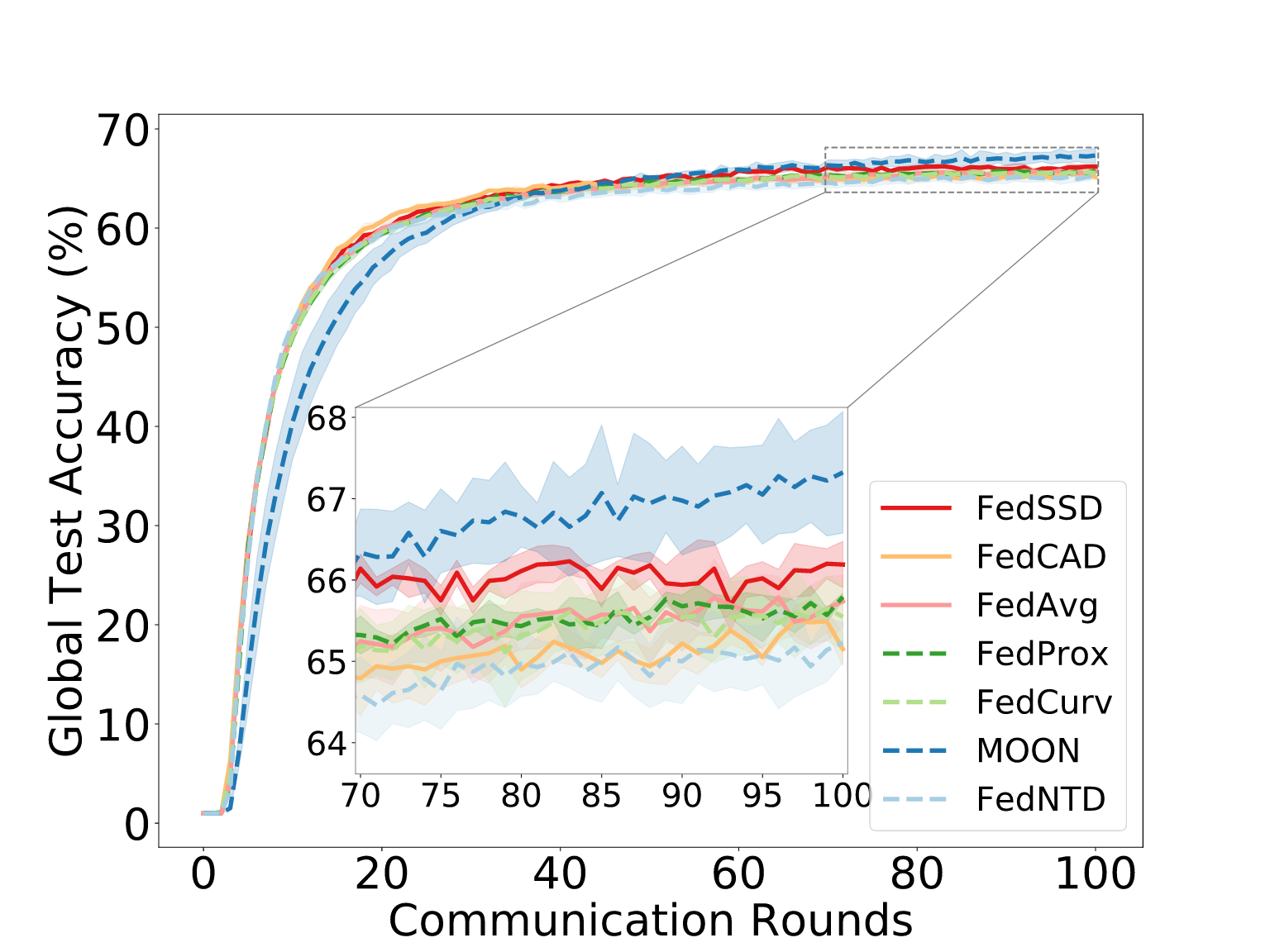} \\
	    \includegraphics[width=1\linewidth]{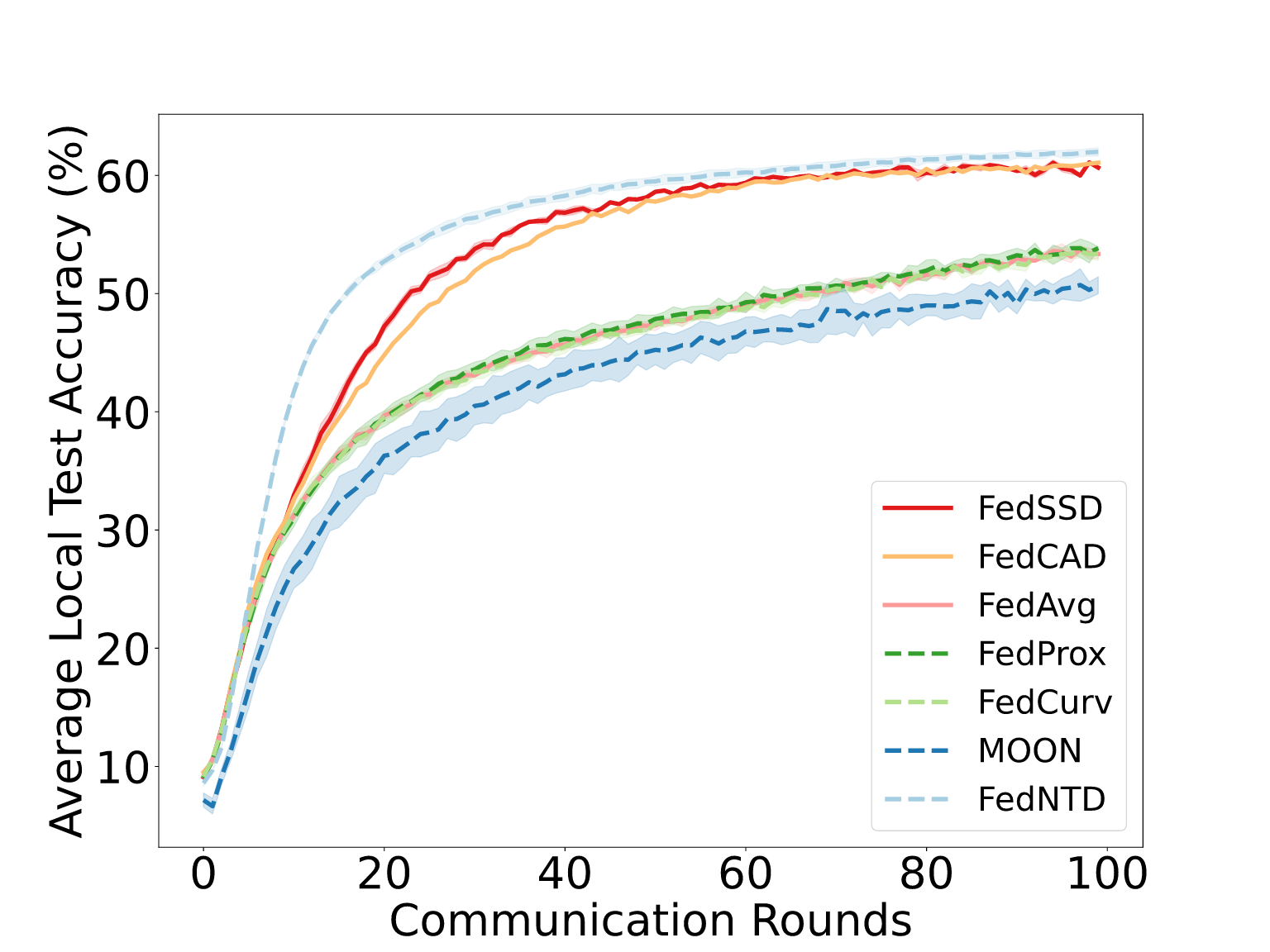}\\

	\end{minipage}
}
	\subfloat[TinyImageNet]{
	\begin{minipage}[b]{0.3\linewidth}
	\includegraphics[width=1\linewidth]{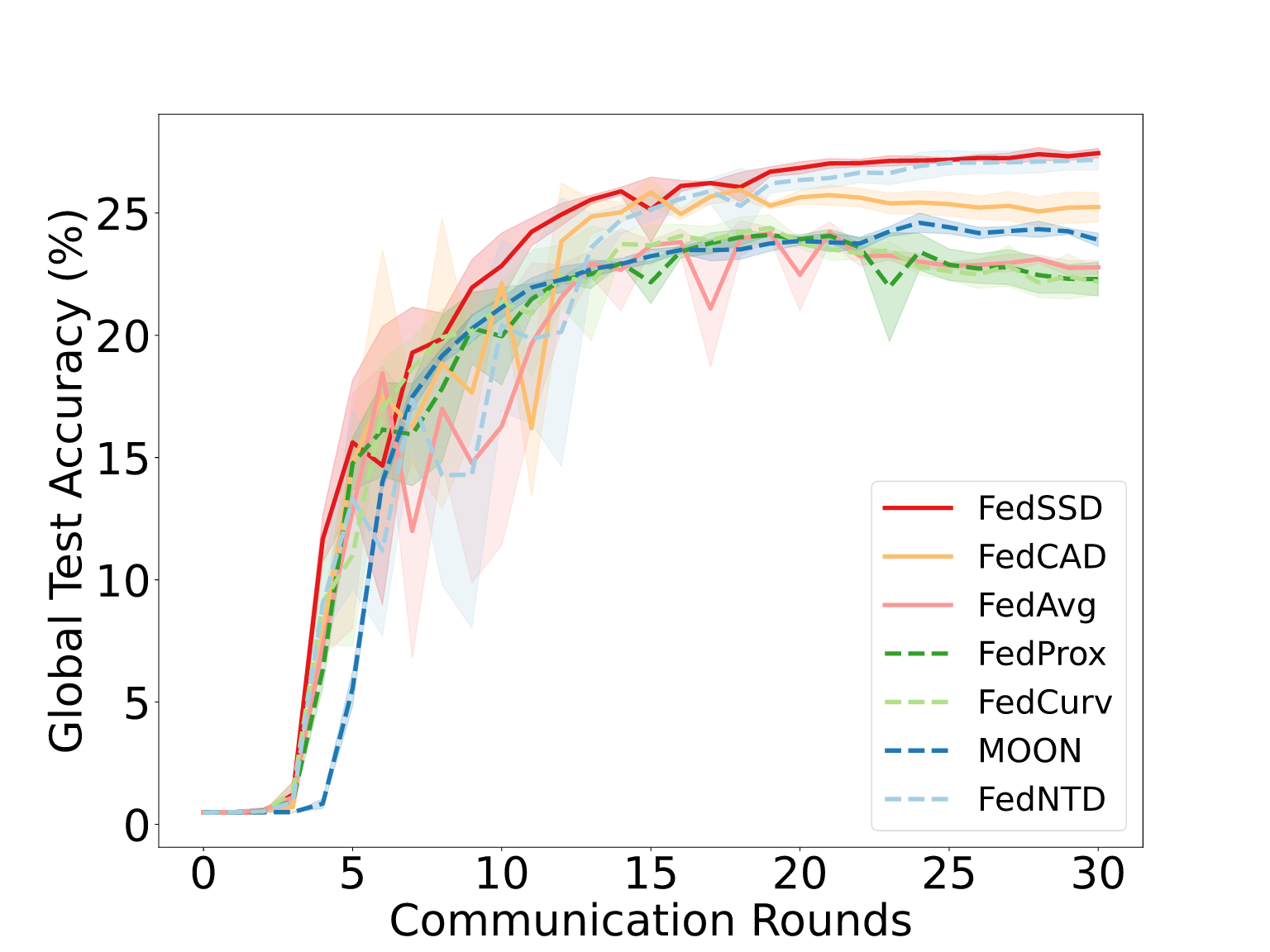} \\
		\includegraphics[width=1\linewidth]{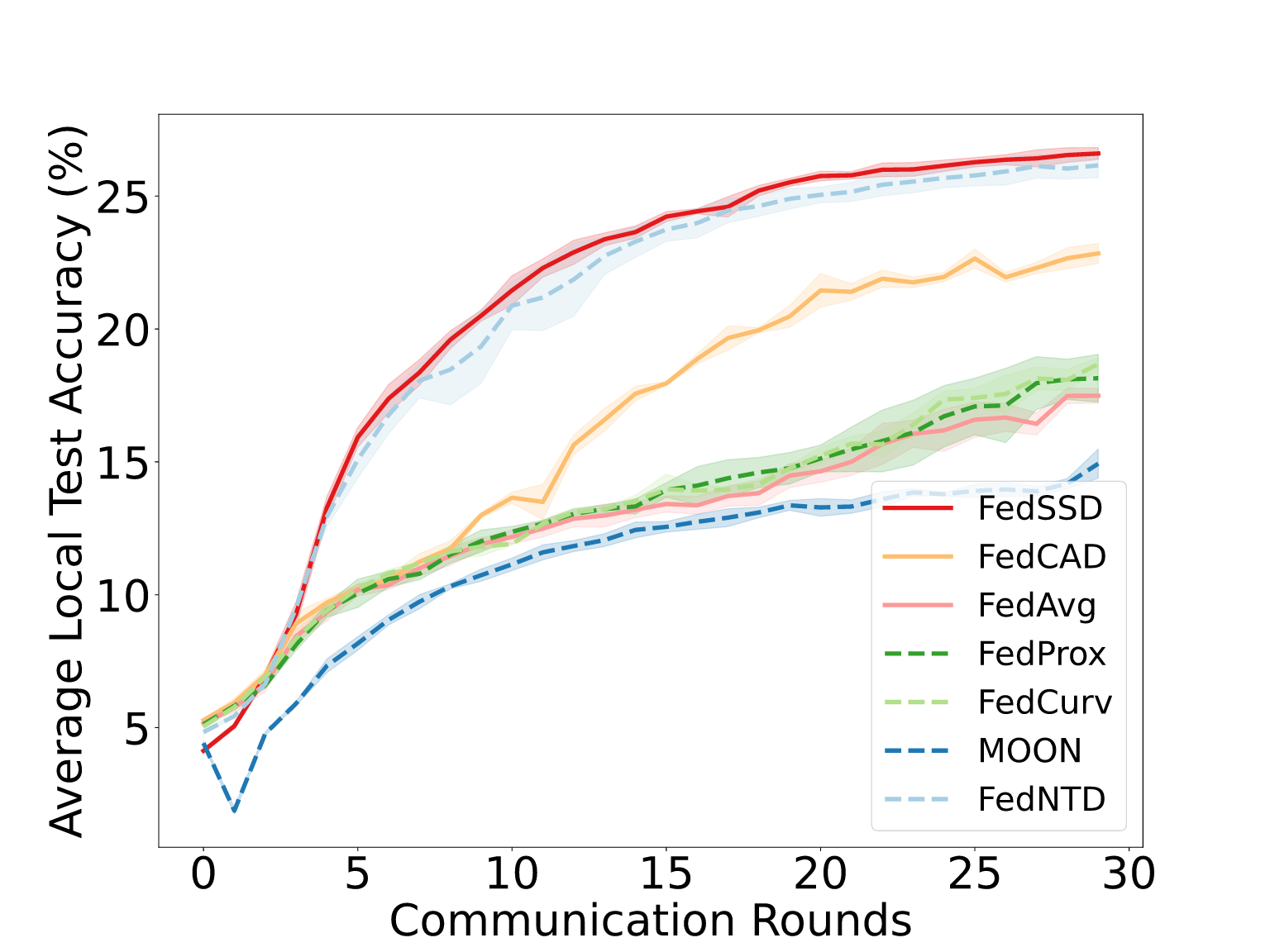}\\

	\end{minipage}
}
\caption{The learning curves on the benchmark datasets with default settings. The global test accuracy is the performance of the global model on the test dataset. The average local test accuracy is the average of the respective performance of the local models on the same test dataset.}
\label{default}
\end{figure*}

%%%%%%%%%%%%%%%%%%%%%%%%%%%%%%%%%  SOTA %%%%%%%%%%%%%%%%%%%%%%%%%%%%%%%%%%%%%%%%%

\section{Experiments} \label{sec:experiments}
In this section, we extensively evaluate our method to demonstrate that critically learning from the global model accelerates the federated learning progress with alleviated knowledge forgetting in the local training phase. Our selective self-distillation strategy, FedSSD, outperforms other methods on non-IID datasets in terms of performance and convergence speed. We design the experiment like \cite{moon}. The implementation details and extended experimental results are provided as follows.

\subsection{Datasets and Experimental Settings}
\noindent \textbf{Datasets. } We evaluate our method on three image datasets: CIFAR10, CIFAR100 \cite{cifar10}, TinyImageNet \cite{le2015tiny}. 
We choose two data partitioning strategies to simulate the non-IID data distribution, which is inspired by \cite{experimental}.
One strategy Quantity-based label imbalance randomly assigns $k$ different labels to each client and we use $\#K=k$ to denote it. The other strategy Distribution-based label imbalance allocates a proportion of the samples of each class to each client according to Dirichlet distribution. Specifically, we sample $p_k \sim Dir(\delta)$ and allocate a proportion $p_{k,i}$ of the instances of class $k$ to client $i$, where $\delta$ is the concentration parameter controlling the uniformity between clients and a bigger $\delta$ indicates a more uniform distribution. With the above partitioning strategy, each client only owns a partial class set. An example of the data distributions among clients is shown in Figure \ref{data_distribution}.
The auxiliary dataset at the server $D_V$ is a small subset of samples of different classes, which can be acquired from the public data. In our experiments, we sample auxiliary data according to the strategy from \cite{ratio_loss}, using only 64 samples for each class.\\
\noindent \textbf{Models. } For CIFAR10, we follow the same network architecture as FedAvg: two $5 \times 5$ convolution layers (the first with 6 channels and the second with 16 channels, each followed by a ReLU activation and $2 \times 2$ max pooling), two fully connected layers with ReLU activation (the first with 120 units and the second with 84 units) and a final softmax output layer. For CIFAR100 and TinyImageNet, we use ResNet50 \cite{resnet} instead. \\
\noindent \textbf{Configurations. } Unless otherwise mentioned, we run 100 rounds on CIFAR10/100 and 30 rounds on TinyImageNet, with 10 clients ($N=10$) and all fixed clients are selected ($C=1$) in each communication round to eliminate the effect of randomness brought by client sampling \cite{fedavg}. We use distribution-based strategy $Dir(\delta=0.5)$ to generate the non-IID data distribution by default. 
In the local training phase, we use the SGD optimizer with initial learning rate 0.01 and momentum 0.9. The local epoch is set to 10 and local batch size is set to 64 by default. We implement FedSSD and the baseline methods with PyTorch \cite{pytorch}, and train our models on RTX 3090 GPU.\\
\noindent \textbf{Baselines. } We compare the performance of FedSSD against recent state-of-the-art (SOTA) FL methods towards solving the non-IID problem, including \textbf{FedAvg}, \textbf{FedProx}, \textbf{FedCurv}, \textbf{MOON}, \textbf{FedNTD} and the previous work \textbf{FedCAD}. The best weight of proximal term $\mu$ in FedProx and the best upper bound $M_{max}$ of FedSSD for CIFAR10, CIFAR100, TinyImageNet are 0.01, 0.001 and 0.001, respectively. The best weight $\lambda$ in FedCurv for CIFAR10, CIFAR100, TinyImageNet are $10^{-3}$, $10^{-4}$ and $10^{-5}$, respectively. We use an additional projection head (2-layer MLP with output dimension 256) to achieve the better performance of MOON and the best weight $\mu$ of model-contrastive loss for CIFAR10, CIFAR100 and TinyImageNet are 5, 1 and 1, respectively.

%%%%%%%%%%%%%%%%%%%%%%%%% Table: SOTA  %%%%%%%%%%%%%%%%%%%%%%%%% 
\begin{table}[htbp]
\caption{The top-1 test accuracy (ACC) after training the target rounds and the number of communication rounds (T) to achieve the same accuracy as running FedAvg. The \textbf{best} and the \underline{second best} values are highlighted.}
\label{tab:accuracy}
\resizebox{\columnwidth}{!}{
\begin{tabular}{lcccccc}
\hline
\multirow{2}{*}{Method} & \multicolumn{2}{c}{CIFAR10} & \multicolumn{2}{c}{CIFAR100} & \multicolumn{2}{c}{TinyImageNet} \\ 
        & ACC $\uparrow$  & T $\downarrow$ & ACC $\uparrow$   & T $\downarrow$ & ACC $\uparrow$ & T $\downarrow$ \\ \hline
FedAvg  & 70.49$\pm 0.23$  & 100          & 65.73$\pm 0.31$ & 100           & 22.78$\pm 0.28$ & 30           \\
FedProx & 72.08$\pm 0.58$ & 55          & 65.79$\pm 0.05$ & 100          & 22.29$\pm 0.69$ & 25           \\
FedCurv & 70.40$\pm 0.74$ & 99          &  65.54$\pm 0.41$ & \_           & 22.18$\pm 0.53$ & 16           \\
MOON  & 71.27$\pm 0.14$ & 62          & \textbf{67.32}$\pm 0.74$  & 60           &  23.89$\pm 0.27$ & 16           \\ 
FedNTD & 70.99$\pm 0.24$ & 51          &  65.23$\pm 0.25$ & \_           &  \underline{27.17}$\pm 0.42$ & 13           \\ \hline
FedCAD & \underline{72.23}$\pm 0.23$ & 35          &  65.15$\pm 0.21$ & \_           &  25.24$\pm 0.26$ & 12           \\
FedSSD  & \textbf{73.38$\pm 0.46$} & \textbf{33}  & \underline{66.19$\pm 0.28$}   & \textbf{48}   & \textbf{27.44$\pm 0.19$} & \textbf{10}           \\ \hline
\end{tabular}}
\end{table}
%%%%%%%%%%%%%%%%%%%%%%%%% Table: SOTA  %%%%%%%%%%%%%%%%%%%%%%%%% 

%%%%%%%%%%%%%%%%%%%%%%%%%%%%%%%%% sample_ratio %%%%%%%%%%%%%%%%%%%%%%%%%%%%%%%%%%%%%%%%%
\begin{figure*} [t]
	\centering
	\subfloat[CIFAR10 ($C=0.1$)]{
	\begin{minipage}[b]{0.24\linewidth}
	\includegraphics[width=1\linewidth]{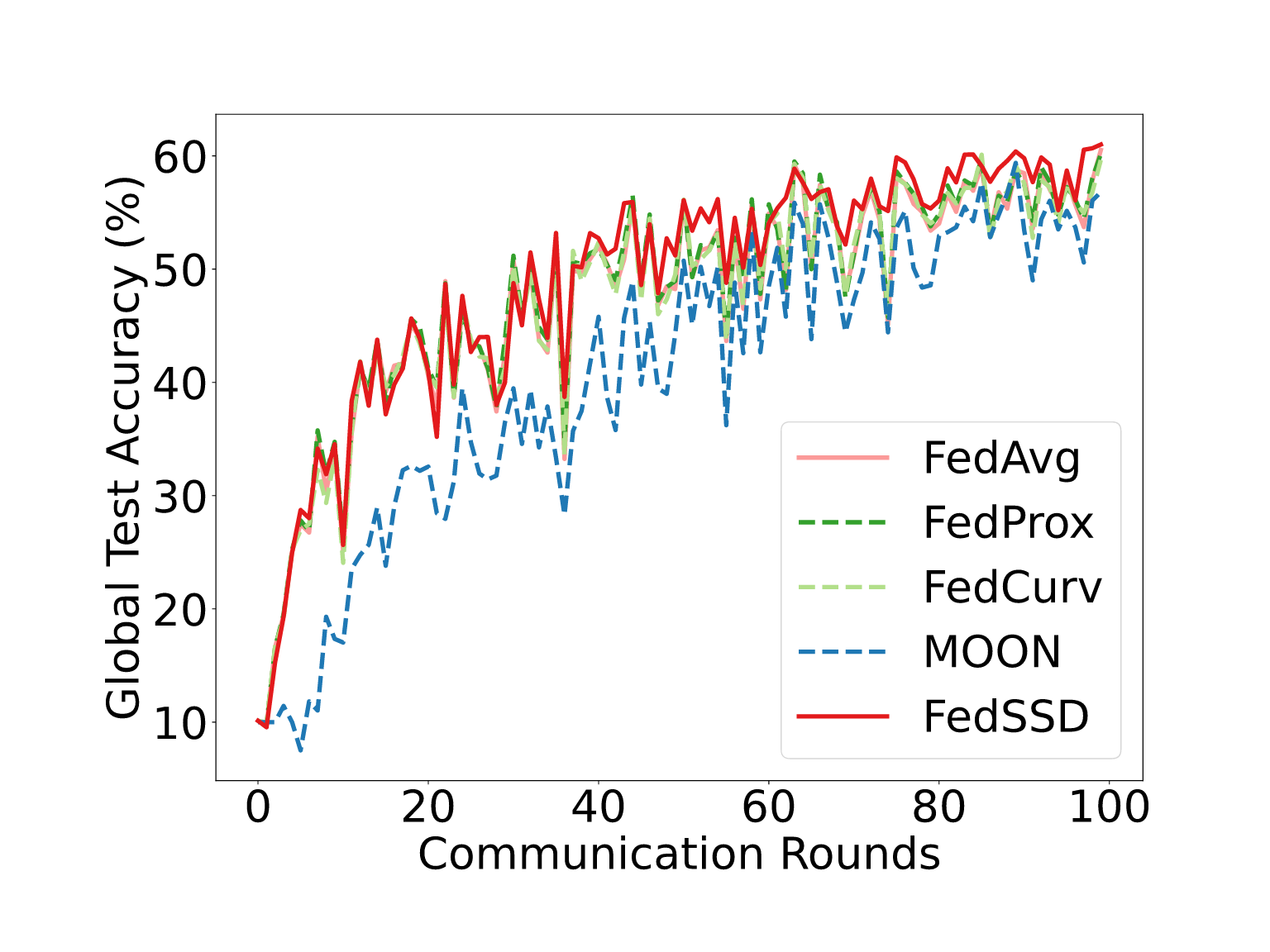}\\
		\includegraphics[width=1\linewidth]{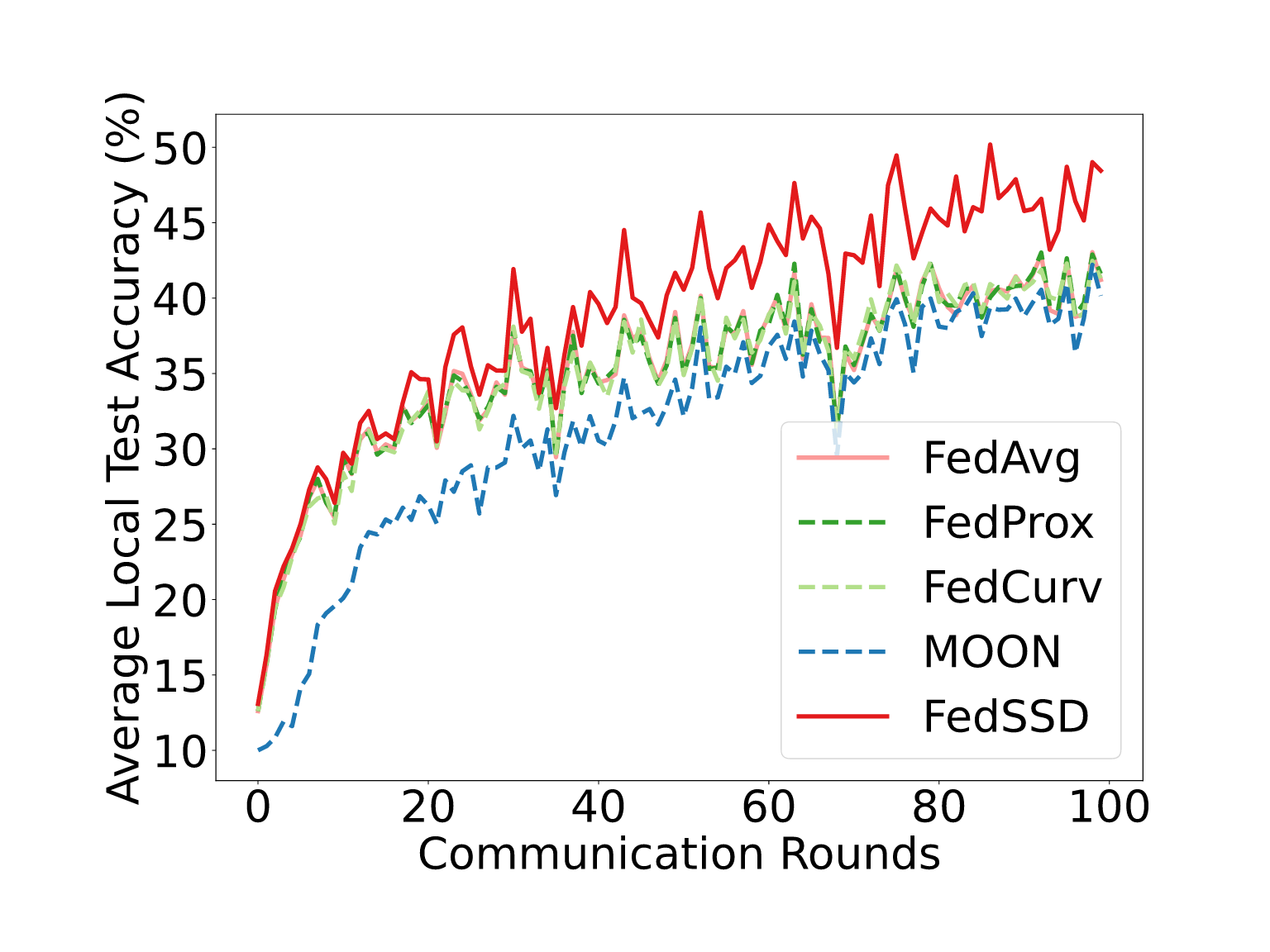}
	\end{minipage}
}
	\subfloat[CIFAR10 ($C=0.2$)]{
	\begin{minipage}[b]{0.24\linewidth}
		\includegraphics[width=1\linewidth]{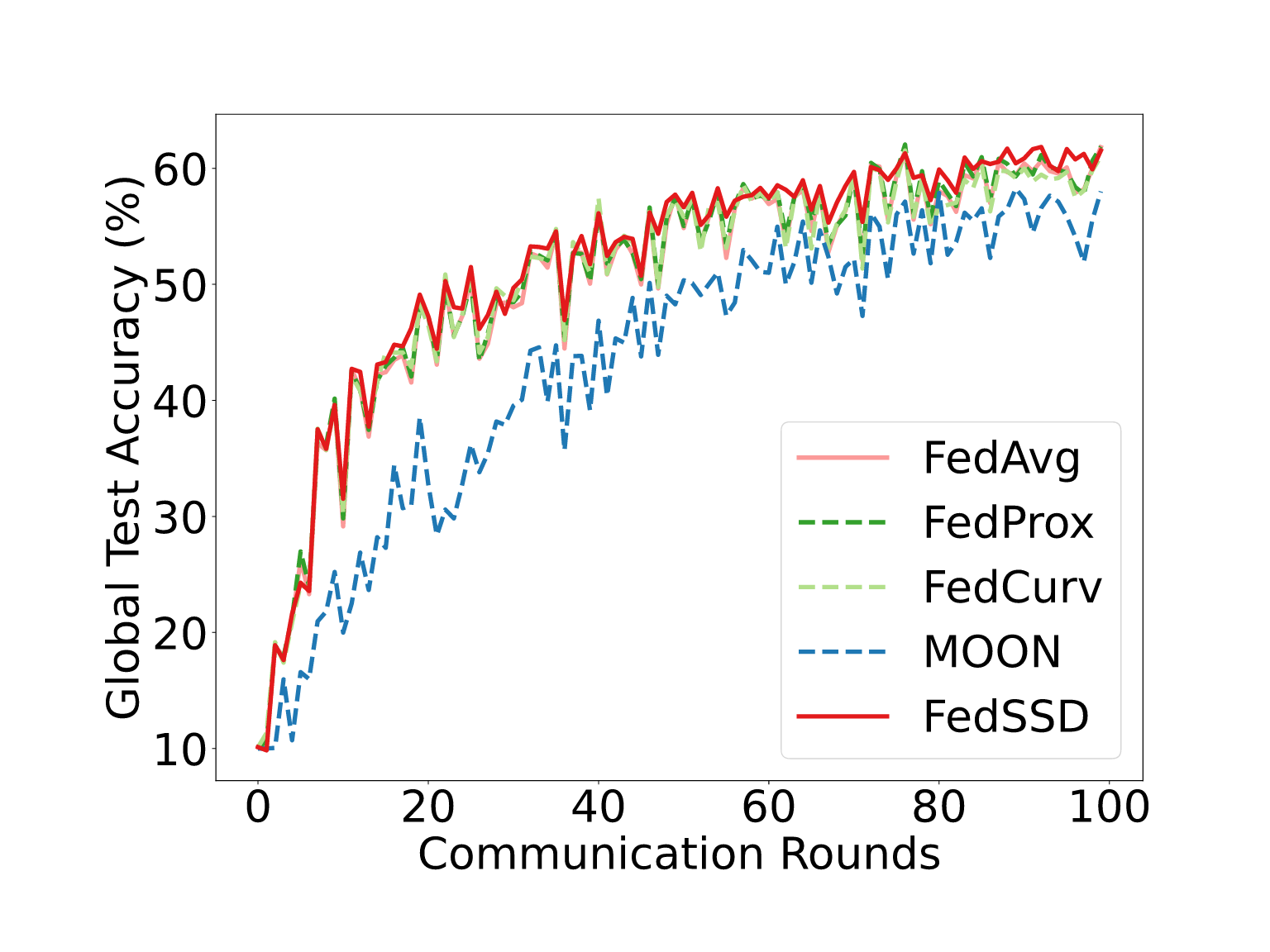} \\
	    \includegraphics[width=1\linewidth]{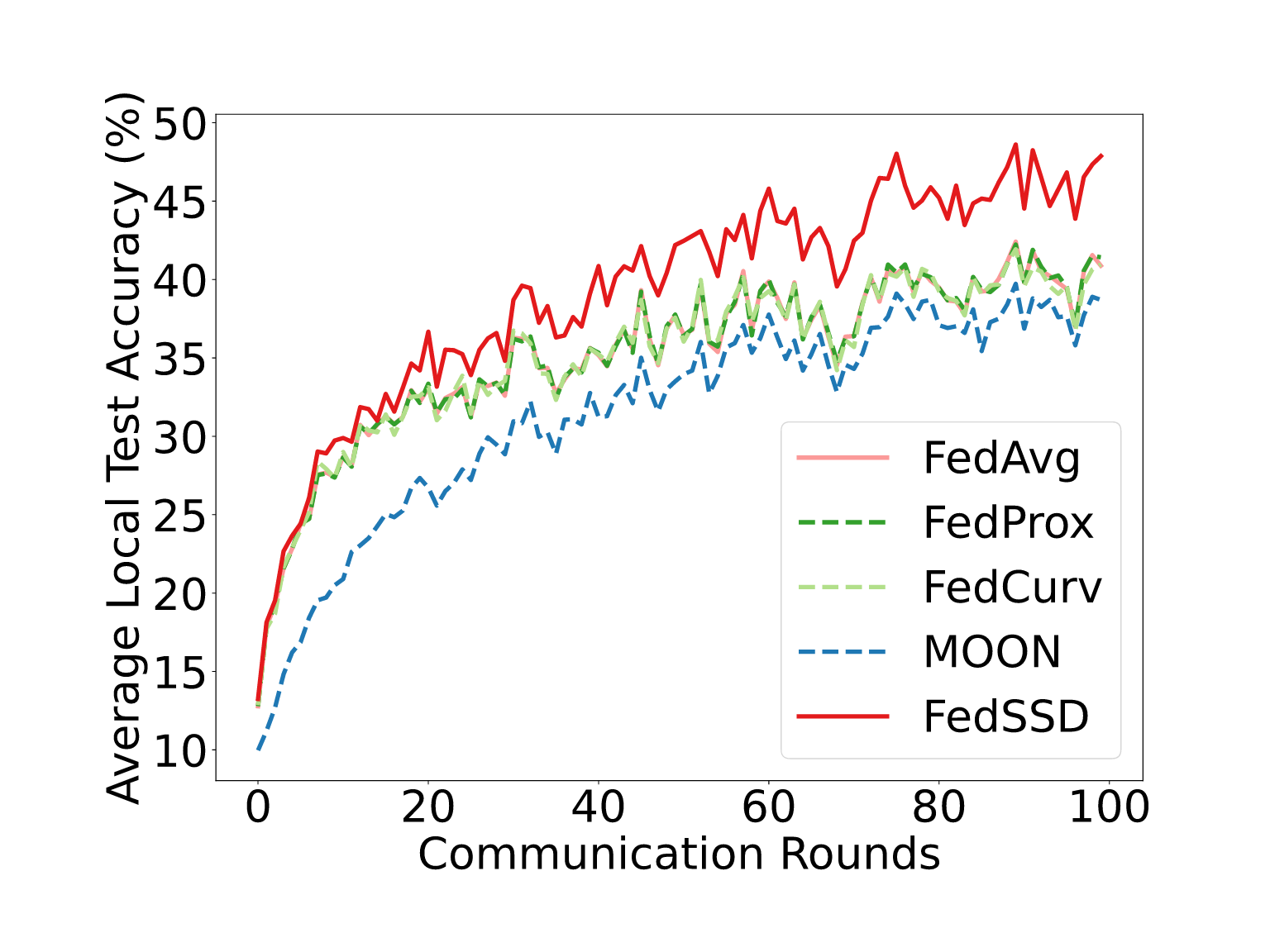}
	\end{minipage}
}
	\subfloat[CIFAR100 ($C=0.1$)]{
	\begin{minipage}[b]{0.24\linewidth}
		\includegraphics[width=1\linewidth]{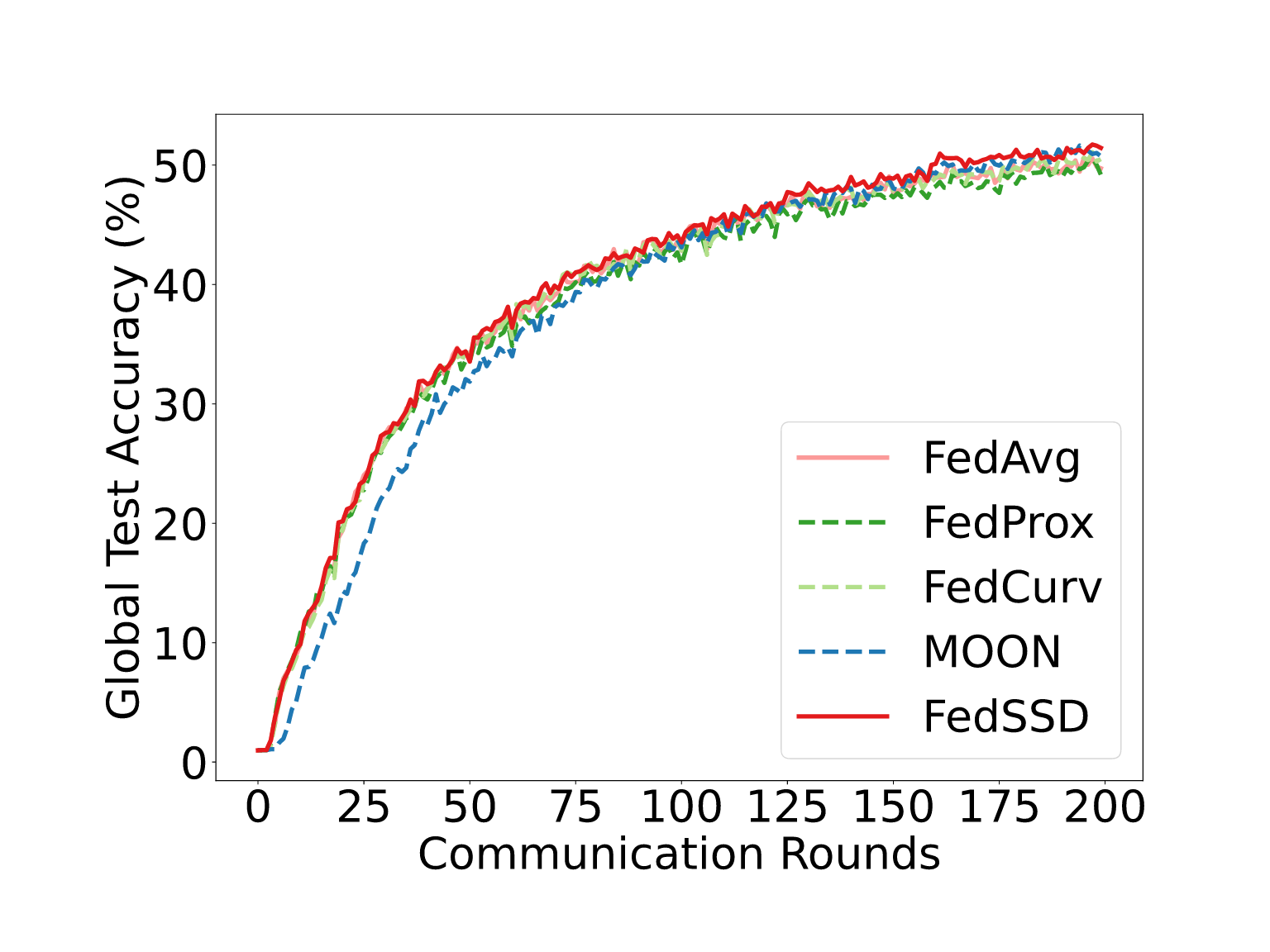} \\
		\includegraphics[width=1\linewidth]{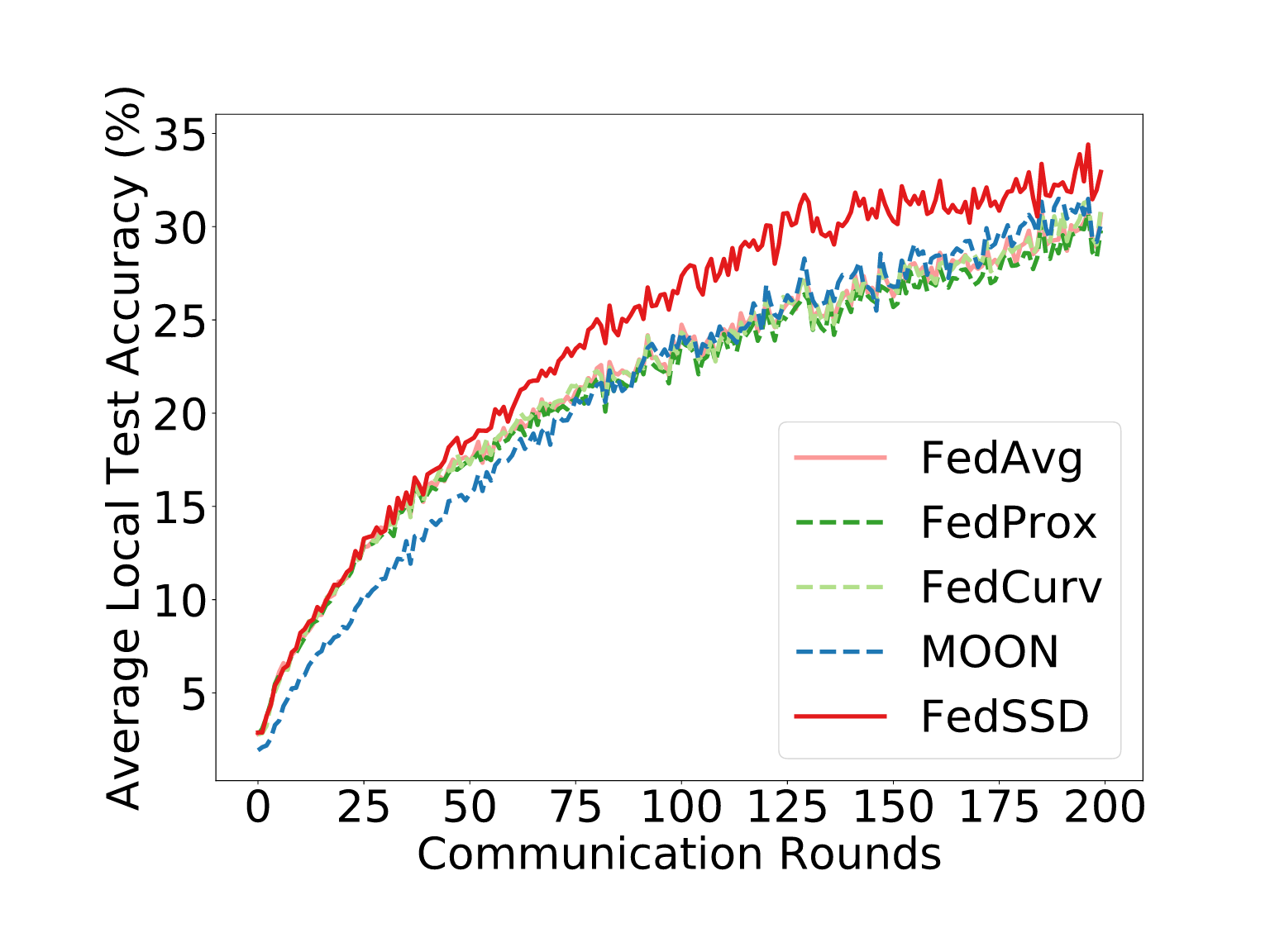}
	\end{minipage}
}
    \subfloat[CIFAR100 ($C=0.2$)]{
	\begin{minipage}[b]{0.24\linewidth}
		\includegraphics[width=1\linewidth]{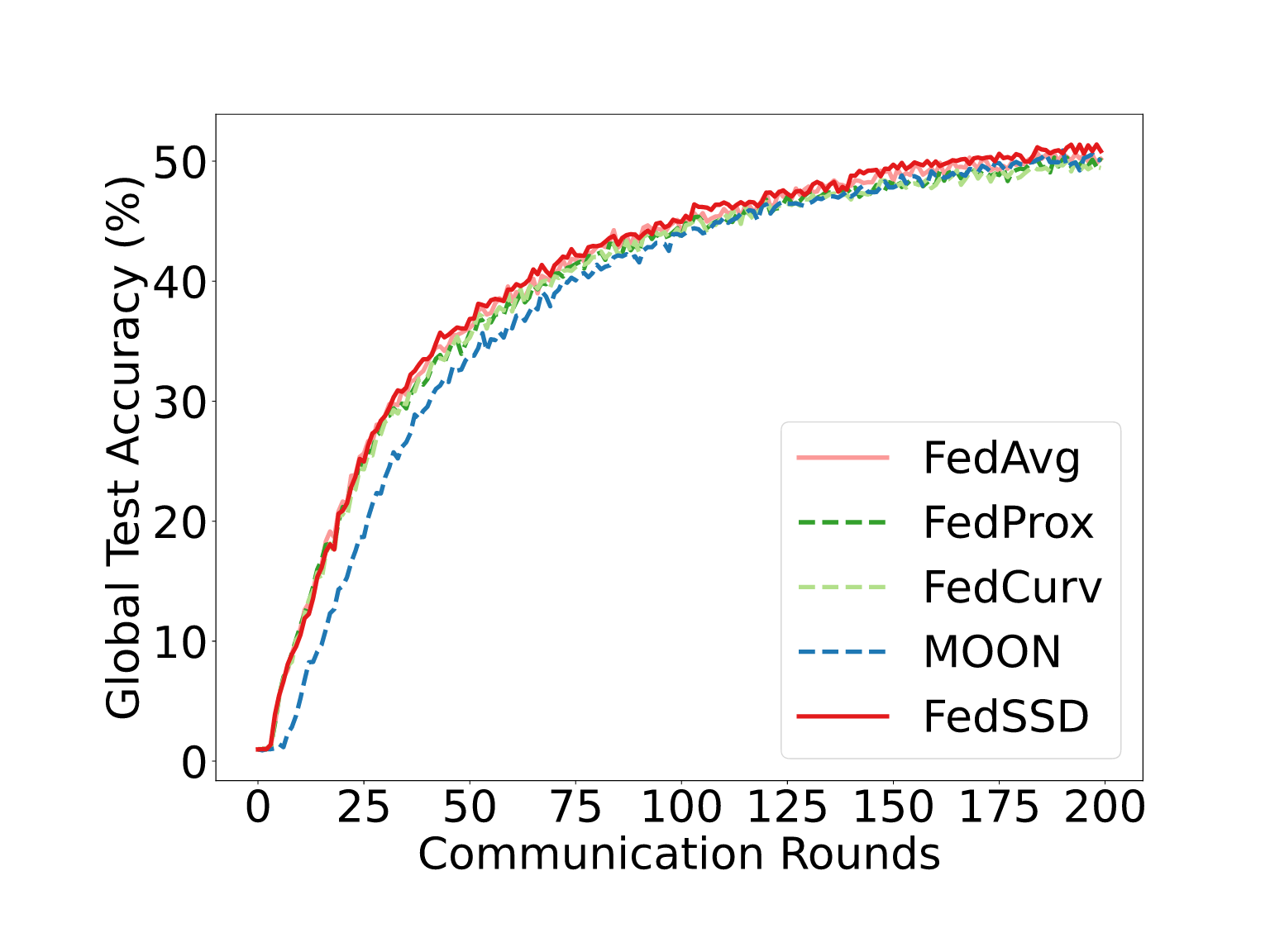} \\
		\includegraphics[width=1\linewidth]{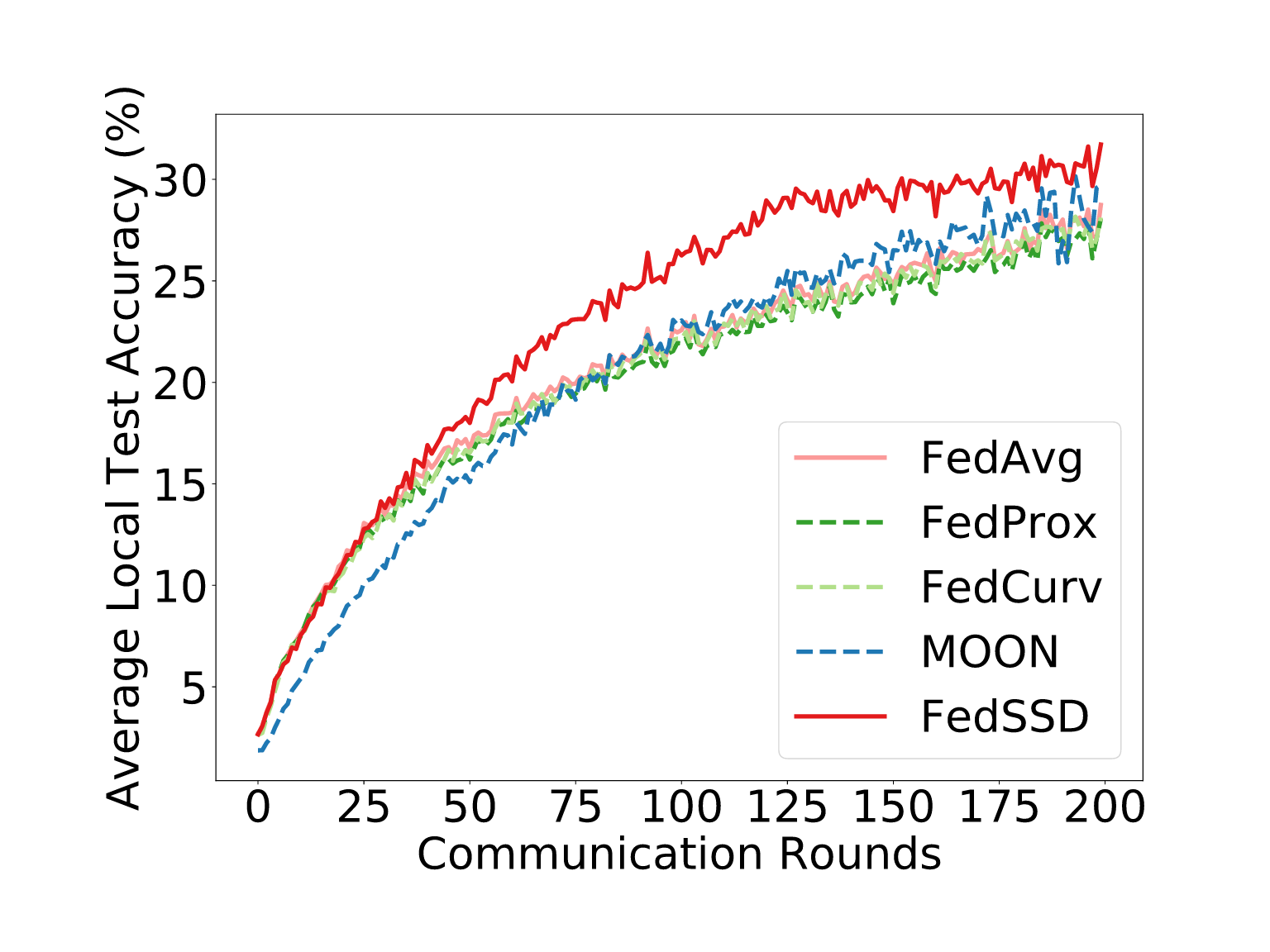}
	\end{minipage}
}
\caption{The learning curves on CIFAR10 and CIFAR100 with different client sample ratios.}
\label{sample_ratio}
\end{figure*}
%%%%%%%%%%%%%%%%%%%%%%%%%%%%%%%%% sample_ratio %%%%%%%%%%%%%%%%%%%%%%%%%%%%%%%%%%%%%%%%%

\subsection{Comparison with the State-Of-The-Arts}
The top-1 test accuracy with the above default settings is shown in Table \ref{tab:accuracy} and the learning curves of the global model and local models are shown in Fig. \ref{default}. For CIFAR10, FedSSD outperformed all the baseline methods in both final model test accuracy and convergence speed (measured in communication rounds), indicating the superiority of FedSSD in dealing with non-IID data when a shallow neural network is used. A similar conclusion can be drawn when training a deeper network on TinyImageNet, showing the robustness of FedSSD on various datasets and model architectures. For CIFAR100, note that although the final global test accuracy of FedSSD is slightly worse than MOON, the average local test accuracy and convergence speed are higher.
In the first row of Fig. \ref{default}, we record the test accuracy of the global model in each communication round. It is obvious that the speed of accuracy improvement in FedSSD is almost the same as FedAvg at the beginning, due to the low credibility of the global model and knowledge distillation loss is almost close to zero. Then, it achieves a better accuracy benefit from the selective knowledge distillation loss, after the global model gains credible knowledge.  

Furthermore, we report the number of communication rounds to reach the same accuracy as running FedAvg for 100 rounds on CIFAR10/100 or 30 rounds on TinyImageNet in Table \ref{tab:accuracy}. The results show that FedSSD significantly outperformed all the baseline methods in the convergence speed (required fewer communication rounds to reach the target accuracy), demonstrating FedSSD accelerates the convergence by alleviating the catastrophic forgetting issue. FedCurv and FedNTD also try to alleviate the forgetting issue, but their performance is not significantly better than ours.

\begin{table}
\caption{The top-1 test accuracy with different levels of data heterogeneity on CIFAR10.}
\centering
\resizebox{\columnwidth}{!}{
\begin{tabular}{lcccccc}
\hline
Method & $\#K=2$ & $\#K=5$ & $\delta=0.1$  & $\delta=0.5$    & $\delta=10$  \\ \hline
FedAvg    & 49.64$\pm 1.64$ & 67.82$\pm 0.28$ & 62.18$\pm 0.54$ & 70.49$\pm 0.23$  & 73.76$\pm 0.27$ \\
FedProx   & 50.31$\pm 1.34$ & 67.39$\pm 0.14$ & 62.61$\pm 0.21$ & 72.08$\pm 0.58$  & 73.29$\pm 0.48$ \\
FedCurv   & 50.37$\pm 1.54$ & 68.15$\pm 0.46$ & 62.09$\pm 0.52$ & 70.40$\pm 0.74$  & 73.49$\pm 0.68$ \\
MOON      & 45.15$\pm 0.75$ & 67.41$\pm 0.13$ & 62.41$\pm 0.26$ & 71.27$\pm 0.14$  & 73.37$\pm 0.11$ \\
FedNTD  & 49.16$\pm 1.92$ & 68.17$\pm 0.31$ & 62.59$\pm 0.25$ & 70.99$\pm 0.24$  & 73.82$\pm 0.22$ \\ \hline
FedCAD  & 50.75$\pm 1.19$ & 68.86$\pm 0.37$ & 62.74$\pm 0.31$ & 72.23$\pm 0.23$  & 74.10$\pm 0.55$ \\
FedSSD    & \textbf{53.37$\pm 1.52$} & \textbf{69.33$\pm 0.47$} & \textbf{62.84}$\pm 0.38$ & \textbf{73.38$\pm 0.46$} & \textbf{74.40$\pm 0.15$}  \\ \hline
\end{tabular}}
\label{tab:data_heterogeneity}
\end{table}

%%%%%%%%%%%%%%%%%%%%%%%%%%%%%%%%% Table: nonIID %%%%%%%%%%%%%%%%%%%%%%%%%%%%%%%%%%%%%%%%%

\subsection{Preservation of the Global Knowledge}
To evaluate the effectiveness of our method in the local training phase, we also record the average test accuracy of the local models on a test dataset obeying the global data distribution. If local models preserve the global knowledge after fitting on the biased local data, the updated local model could be generalized well on the uniform test data distribution. As shown in the second row of Fig. \ref{default}, the improved speed of local test accuracy in FedSSD is almost the same as FedAvg at the beginning. Since the global model is far from convergence and the credibility of the global model is low. Then, FedSSD achieves better accuracy than FedAvg thanks to the distillation loss after about 10 rounds, which helps local models to preserve the global knowledge and mitigate the catastrophic forgetting during local training. The result demonstrates that the selective self-distillation mechanism makes local models more generalized on the global data distribution.

\subsection{Sensitivity Analysis}
\noindent \textbf{Effects of data heterogeneity. }
To evaluate the robustness of our method under different data heterogeneity levels, we use quantity-based $\#K=k$ and Dirichlet distribution-based $Dir(\delta)$ strategies to partition CIFAR10 and CIFAR100. A bigger $k$ and $\delta$ indicate a more uniform distribution.
Results in Table \ref{tab:data_heterogeneity} show that our method consistently outperforms other baselines across different heterogeneity levels, and the trend of which is more pronounced at higher data heterogeneity. Under the quantity-based data heterogeneity ($\#K=2,5$), the performance of MOON is worse than FedAvg, while FedSSD still outperforms others. Although our method outperforms FedCurv only 1\% accuracy, the communication costs of FedCurv increase to 2 times compared with our method, due to our credibility matrix is simpler than the Fisher information.
The experiments demonstrate the robustness and effectiveness of FedSSD under different heterogeneity levels. \\

\noindent \textbf{Effects of clients participation ratio. }
To evaluate the scalability of our method, we design the experiments with a different numbers of participating clients at each communication round on CIFAR10 and CIFAR100. Specifically, we partition CIFAR10 and CIFAR100 training datasets into 100 clients and randomly sample 10, 20 clients to participate in the training during each round. The results are shown in Fig. \ref{sample_ratio}, FedSSD stably outperforms other methods even with fewer participants in each round, which indicates FedSSD has strong robustness to different client participation degrees. \\

%%%%%%%%%%%%%%%%%%%%%%%%%%%%%%%%% Figure: epochs  %%%%%%%%%%%%%%%%%%%%%%%%%%%%%%%%%%%%%%%%%
\begin{figure}[b]
\centering
\includegraphics[width=0.9\linewidth]{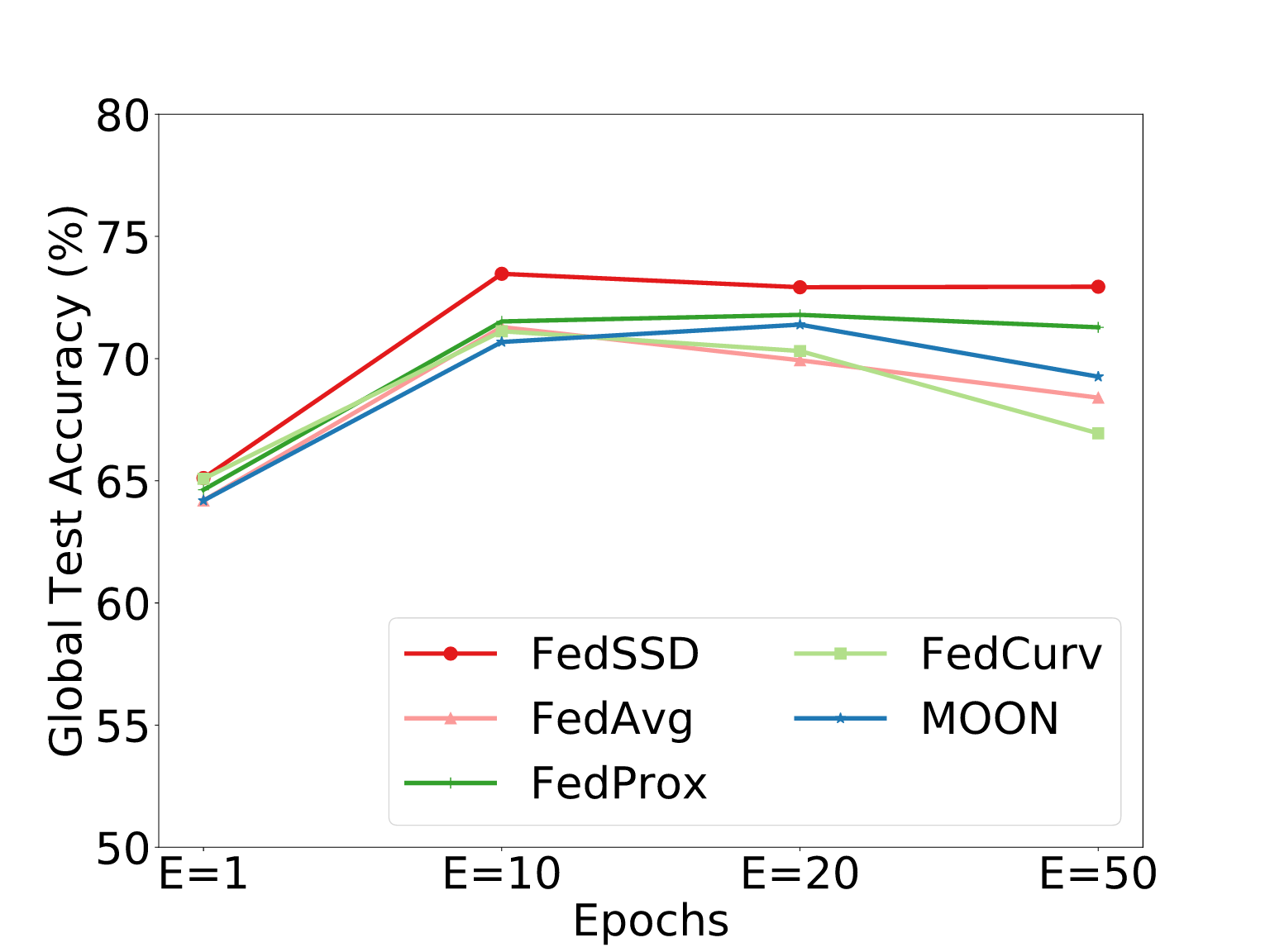}
\caption{Visualized the final global model's performance with different number of local epochs on CIFAR10.}
\label{epochs}
\end{figure}
%%%%%%%%%%%%%%%%%%%%%%%%%%%%%%%%% Figure: epochs  %%%%%%%%%%%%%%%%%%%%%%%%%%%%%%%%%%%%%%%%%

\noindent \textbf{Effects of local update epochs. }
Aggregating local models at different frequencies may affect the learning performance, since less frequent communication will further enhance the drift in the local training phase. 
We conduct the experiments to study the effect of local epochs on the performance of the final global model and the results are shown in Fig. \ref{epochs}. Intuitively, a small $E$ may increase the communication burden and a large $E$ may result in a low convergence rate. When the local epoch $E=1$, all methods have a relatively close test accuracy due to each client communicates with others frequently and there is no "Client drift". With more local epochs are used, the client local update drifts more and the discrepancy between the local model and global model becomes larger. Both FedAvg and MOON suffer from performance drops. Nevertheless, due to our methods selectively self-distills the global knowledge into local models, FedSSD significantly outperforms other methods and shows robustness to large drift caused by more local update epochs.\\

\noindent \textbf{Ablation Study.}
We further conduct an ablation study to investigate key properties of FedSSD. We replace our distillation loss with KL-divergence distillation loss and MSE distillation loss respectively. Both two losses are controlled by a constant coefficient rather than our adaptive weights. We uniformly denote the constant coefficient and the upper bound of selective distillation loss $M_{max}$ as $\alpha$.
As shown in Table \ref{tab:Ablation}, selective distillation achieves best performance under different values of $\alpha$.

\begin{table}[]
\caption{The top-1 test accuracy with different loss types on CIFAR10.}
\label{tab:Ablation}
\centering
\begin{tabular}{llll}
\hline
\multicolumn{1}{c}{Loss Type} & \multicolumn{1}{c}{$\alpha=0.01$} & \multicolumn{1}{c}{$\alpha=0.1$} & \multicolumn{1}{c}{$\alpha=0.5$} \\ \hline
KL  & 71.27 & 71.30 & 70.61 \\
MSE & 72.66  & 71.44 & 69.51 \\
SSD & 73.47 & 71.57 & 71.38 \\ \hline
\end{tabular}
\end{table}

\section{Conclusion}
In this work, we propose a novel federated learning algorithm with selective self-distillation (FedSSD), to overcome the forgetting issue in the local training phase. We observe that the global model learns a better representation than local models but is not reliable for distilling in every case. FedSSD critically distills the knowledge of the global model into local models by measuring the credibility of samples and class channels of logits, which helps local models to preserve the global knowledge and meanwhile learns the knowledge from the local data.
The effectiveness of the proposed method has been comprehensively analyzed from both theoretical and experimental perspectives. FedSSD can be used for non-vision problems in the future because it does not require image inputs.

\ifCLASSOPTIONcompsoc
  \section*{Acknowledgments}
\else
  \section*{Acknowledgment}
\fi
This work is supported by the National Key Research \& Development Plan of China No.2021YFC2501202, Natural Science Foundation of China No.61972383 and No. 61902377, Beijing Municipal Science \& Technology Commission No.Z211100002121171, Youth Innovation Promotion Association CAS, Science and Technology Service Network Initiative, Chinese Academy of Sciences No. KFJ-STS-QYZD-2021-11-001
\ifCLASSOPTIONcaptionsoff
  \newpage
\fi

\bibliographystyle{IEEEtran}
\bibliography{references}{}

\end{document}